\documentclass{article}

\usepackage[nonatbib, final]{neurips_2021}

\usepackage[utf8]{inputenc} 
\usepackage[T1]{fontenc}    
\usepackage{hyperref}       
\usepackage{url}            
\usepackage{booktabs}       
\usepackage{amsfonts}       
\usepackage{nicefrac}       
\usepackage{microtype}      
\usepackage{xcolor}         

\usepackage{tabulary,multirow,xspace}
\usepackage{fixmath,mathtools,nicefrac,mmstyle}
\usepackage{cite}
\usepackage{subcaption}
\usepackage{caption}
\usepackage{float}
\usepackage{wrapfig} 
\usepackage[misc]{ifsym} 

\newcommand{\methodname}{K-Net}
\newcommand{\shortname}{K-Net}
\newcommand{\myparagraph}[1]{{\noindent\bf #1}}

\newlength\savewidth\newcommand\shline{\noalign{\global\savewidth\arrayrulewidth
  \global\arrayrulewidth 1pt}\hline\noalign{\global\arrayrulewidth\savewidth}}
\newcolumntype{x}[1]{>{\centering\arraybackslash}p{#1pt}}
\newcommand{\tablestyle}[2]{\setlength{\tabcolsep}{#1}\renewcommand{\arraystretch}{#2}\centering\small}

\newcommand{\things}{\tss{Th}\xspace}
\newcommand{\stuff}{\tss{St}\xspace}
\newcommand{\app}{\raise.17ex\hbox{$\scriptstyle\sim$}}
\newcommand{\tss}[1]{\textsuperscript{#1}}

\newcommand{\dt}[1]{\fontsize{8pt}{.1em}\selectfont \emph{#1}}
\newcommand{\bd}[1]{\textbf{#1}}

\title{K-Net: Towards Unified Image Segmentation}

\author{%
  Wenwei Zhang$^1$ \quad Jiangmiao Pang$^{2,4}$ \quad Kai Chen$^{3,4}$ \quad Chen Change Loy$^{1\textrm{\Letter}}$ \\
  $^1$S-Lab, Nanyang Technological University \\
  $^2$CUHK-SenseTime Joint Lab, the Chinese University of Hong Kong\\
  $^3$SenseTime Research \quad $^4$Shanghai AI Laboratory\\
  \texttt{\{wenwei001, ccloy\}@ntu.edu.sg} \quad \texttt{pangjiangmiao@gmail.com} \\
  \quad \texttt{chenkai@sensetime.com}\\
}

\begin{document}

\maketitle

\begin{abstract}
Semantic, instance, and panoptic segmentations have been addressed using different and specialized frameworks despite their underlying connections.
This paper presents a unified, simple, and effective framework for these essentially similar tasks.
The framework, named \textbf{\methodname}, segments both instances and semantic categories consistently by a group of learnable \emph{kernels}, where
each kernel is responsible for generating a mask for either a potential instance or a stuff class.
To remedy the difficulties of distinguishing various instances,
we propose a kernel update strategy that enables each kernel dynamic and conditional on its meaningful group in the input image.
\methodname~can be trained in an end-to-end manner with bipartite matching, and its training and inference are naturally NMS-free and box-free.
Without bells and whistles, K-Net surpasses all previous published state-of-the-art single-model results of panoptic segmentation on MS COCO \texttt{test-dev} split and semantic segmentation on ADE20K val split with \bd{55.2\%} PQ and \bd{54.3\%} mIoU, respectively.
Its instance segmentation performance is also on par with Cascade Mask R-CNN on MS COCO with 60\%-90\% faster inference speeds. Code and models will be released at \url{https://github.com/ZwwWayne/K-Net/}.

\end{abstract}

\section{Introduction}\label{sec:Introduction}\vspace{-1mm}

Image segmentation aims at finding groups of coherent pixels~\cite{szeliski2010computer}. 
There are different notions in groups, such as semantic categories (\eg, car, dog, cat) or instances (\eg, objects that coexist in the same image).
Based on the different segmentation targets, the tasks are termed differently, \ie, semantic and instance segmentation, respectively.
There are also pioneer attempts~\cite{holistic_scene_understanding, panoptic, Gould_2009, tu_2005} to joint the two segmentation tasks for more comprehensive scene understanding.

\begin{figure}[t]
	\centering\includegraphics[width=\linewidth]{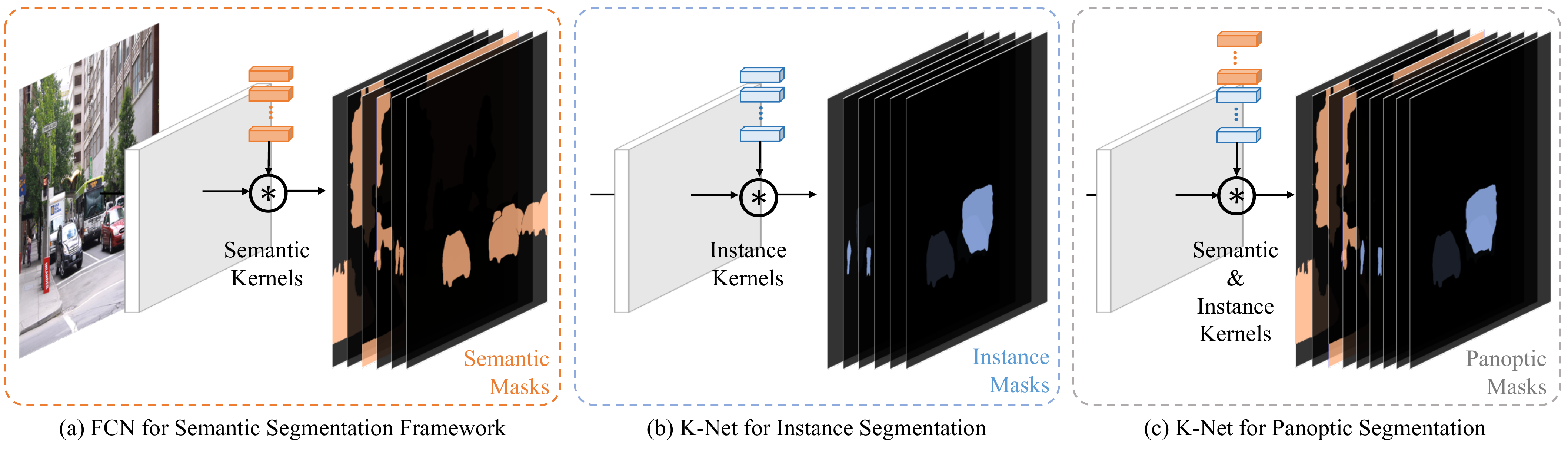}
	\vspace{-15pt}
	\caption{\small{
			Semantic segmentation (a), instance (b), and panoptic segmentation (c) tasks are unified by a common framework in this paper.
			In conventional semantic segmentation methods, each convolutional kernel corresponds to a semantic class.
			Our framework extends this notion to make each kernel corresponds to either a potential instance or a semantic class.
		}}\label{fig:teaser}
	\vspace{-15pt}
\end{figure}

Grouping pixels according to semantic categories can be formulated as a dense classification problem.
As shown in Fig.~\ref{fig:teaser}-(a), recent methods directly learn a set of convolutional kernels (namely \emph{semantic kernels} in this paper) of pre-defined categories and use them to classify pixels~\cite{FCN} or regions~\cite{mask_rcnn}. Such a framework is elegant and straightforward.
However, extending this notion to instance segmentation is non-trivial given the varying number of instances across images.
Consequently, instance segmentation is tackled by more complicated frameworks with additional steps such as object detection~\cite{mask_rcnn} or embedding generation~\cite{associative_embedding}.
These methods rely on extra components, which must guarantee the accuracy of extra components to a reasonable extent, or demand complex post-processing such as Non-Maximum Suppression (NMS) and pixel grouping.
Recent approaches~\cite{solov2,tian2020conditional,panoptic_fcn} generate kernels from dense feature grids and then select kernels for segmentation to simplify the frameworks.
Nonetheless, since they build upon dense grids to enumerate and select kernels,
these methods still rely on hand-crafted post-processing to eliminate masks or kernels of duplicated instances.

In this paper, we make the first attempt to formulate a unified and effective framework to bridge the seemingly different image segmentation tasks (semantic, instance, and panoptic) through the notion of \textit{kernels}.
Our method is dubbed as \methodname~(`K' stands for kernels). It begins with a set of convolutional kernels that are randomly initialized, and learns the kernels in accordance to the segmentation targets at hand, namely, \emph{semantic kernels} for semantic categories and \emph{instance kernels} for instance identities (Fig.~\ref{fig:teaser}-(b)).
A simple combination of semantic kernels and instance kernels allows panoptic segmentation naturally (Fig.~\ref{fig:teaser}-(c)).
In the forward pass, the kernels perform convolution on the image features to obtain the corresponding segmentation predictions.

The versatility and simplicity of \methodname~are made possible through two designs.
First, we formulate \methodname~so that it dynamically updates the kernels to make them conditional to their activations on the image.
Such a content-aware mechanism is crucial to ensure that each kernel, especially an instance kernel, responds accurately to varying objects in an image.
Through applying this adaptive kernel update strategy iteratively, \methodname~significantly improves the discriminative ability of the kernels and boosts the final segmentation performance. It is noteworthy that this strategy universally applies to kernels for all the segmentation tasks.

Second, inspired by recent advances in object detection~\cite{DETR}, we adopt the bipartite matching strategy~\cite{set_prediction_loss} to assign learning targets for each kernel. 
This training approach is advantageous to conventional training strategies~\cite{ren2015faster, lin2017_focal} as it builds a one-to-one mapping between kernels and instances in an image. It thus resolves the problem of dealing with a varying number of instances in an image. In addition, it is purely mask-driven without involving boxes. Hence, \shortname~is naturally NMS-free and box-free, which is appealing to real-time applications.

To show the effectiveness of the proposed unified framework on different segmentation tasks, we conduct extensive experiments on COCO dataset~\cite{lin2014coco} for panoptic and instance segmentation, and ADE20K dataset~\cite{ADE20K} for semantic segmentation.
Without bells and whistles, \shortname~surpasses all previous state-of-the-art single-model results on panoptic (\bd{54.6\%} PQ) and semantic segmentation benchmarks (\bd{54.3\%} mIoU) and achieves competitive performance compared to the more expensive Cascade Mask R-CNN~\cite{cascade_rcnn}.
We further analyze the learned kernels and find that instance kernels incline to specialize on objects at specific locations of similar sizes.
%

\vspace{-1mm}\section{Related Work}\vspace{-1mm}
\myparagraph{Semantic Segmentation.}
Contemporary semantic segmentation approaches typically build upon a fully convolutional network (FCN)~\cite{FCN} and treat the task as a dense classification problem.
Based on this framework, many studies focus on enhancing the feature representation through dilated convolution~\cite{deeplab, deeplabv3, deeplabv3plus},
pyramid pooling~\cite{PSPNet, upernet}, context representations~\cite{OCRNet, encnet}, and attention mechanisms~\cite{psanet, DNL, emanet}.
Recently, SETR~\cite{SETR} reformulates the task as a sequence-to-sequence prediction task by using a vision transformer~\cite{VIT}.
Despite the different model architectures, the approaches above share the common notion of making predictions via static semantic kernels.
Differently, the proposed \methodname~makes the kernels dynamic and conditional on their activations in the image.

\myparagraph{Instance Segmentation.}
%
There are two representative frameworks for instance segmentation -- `top-down' and `bottom-up' approaches.
`Top-down' approaches~\cite{mask_rcnn, Dai_2016, FCIS, YOLACT, BCNet} first detect accurate bounding boxes and generate a mask for each box.
Mask R-CNN~\cite{mask_rcnn} simplifies this pipeline by directly adding a FCN~\cite{FCN} in Faster R-CNN~\cite{ren2015faster}.
Extensions of this framework add a mask scoring branch~\cite{ms_rcnn} or adopt a cascade structure~\cite{cascade_rcnn, htc}.
`Bottom-up' methods~\cite{deep_watershed, associative_embedding, NevenBPG19, instance_cut} first perform semantic segmentation then group pixels into different instances.
These methods usually require a grouping process, and their performance often appears inferior to `top-down' approaches in popular benchmarks~\cite{lin2014coco, Cityscapes}.
Unlike all these works, \methodname~performs segmentation and instance separation simultaneously by constraining each kernel to predict one mask at a time for one object.
Therefore, \methodname~needs neither bounding box detection nor grouping process.
It focuses on refining kernels rather than refining bounding boxes, different from previous cascade methods~\cite{cascade_rcnn, htc}.

Recent attempts~\cite{polarmask, chen2019tensormask, solo, solov2} perform instance segmentation in one stage without involving detection nor embedding generation.
These methods apply dense mask prediction using dense sliding windows~\cite{chen2019tensormask} or dense grids~\cite{solo}.
Some studies explore polar~\cite{polarmask} representation, contour~\cite{deep_snake}, and explicit shape representation~\cite{ese_seg} of instance masks.
These methods all rely on NMS to eliminate duplicated instance masks, which hinders end-to-end training. The heuristic process is also unfavorable for real-time applications.
Instance kernels in \methodname~are trained in an end-to-end manner with bipartite matching and set prediction loss, thus, our methods does not need NMS.

\myparagraph{Panoptic Segmentation.} Panoptic segmentation~\cite{panoptic} combines instance and semantic segmentation to provide a richer understanding of the scene.
Different strategies have been proposed to cope with the instance segmentation task.
Mainstream frameworks add a semantic segmentation branch~\cite{panoptic_fpn, UPSNet, unifying_CVPR2020, solov2} on an instance segmentation framework
or adopt different pixel grouping strategies~\cite{panoptic_deeplab, deeperlab} based on a semantic segmentation method.
Recently, DETR~\cite{DETR} tries to simplify the framework by transformer~\cite{transformer} but need to predict boxes around both stuff and things classes in training for assigning learning targets.
These methods either need object detection or embedding generation to separate instances, which does not reconcile the instance and semantic segmentation in a unified framework.
By contrast, \methodname~partitions an image into semantic regions by semantic kernels and object instances by instance kernels through a unified perspective of kernels.

Concurrent to \methodname, some recent attempts~\cite{max_deeplab, cheng2021maskformer, panoptic_segformer} apply Transformer~\cite{transformer} for panoptic segmentation.
MaskFormer reformulates semantic segmentation as a mask classification task, which is commonly adopted in instance-level segmentation.
From an inverse perspective, K-Net tries to simplify instance and panoptic segmentation by letting a kernel to predict the mask of only one instance or a semantic category, which is the essential design in semantic segmentation.
In contrast to \methodname~that directly uses learned kernels to predict masks and progressively refines the masks and kernels,
MaX-DeepLab~\cite{max_deeplab} and MaskFormer~\cite{cheng2021maskformer} rely on queries and Transformer~\cite{transformer} to produce dynamic kernels for the final mask prediction.

\myparagraph{Dynamic Kernels.} Convolution kernels are usually static, \ie, agnostic to the inputs, and thus have limited representation ability.
Previous works~\cite{dcn, dcnv2, STN, DFN, deformable_kernel} explore different kinds of dynamic kernels to improve the flexibility and performance of models.
Some semantic segmentation methods apply dynamic kernels to improve the model representation with enlarged receptive fields~\cite{dynamic_filtering_eccv2018} or multi-scales contexts~\cite{DMNet}.
Differently, \methodname~uses dynamic kernels to improve the discriminative capability of the segmentation kernels more so than the input features of kernels.

Recent studies apply dynamic kernels to generate instance\cite{tian2020conditional, solov2} or panoptic~\cite{panoptic_fcn} segmentation predictions directly.
Because these methods generate kernels from dense feature maps, enumerate kernels of each position, and filter out kernels of background regions,
they either still rely on NMS~\cite{tian2020conditional, solov2} or need extra kernel fusion~\cite{panoptic_fcn} to eliminate kernels or masks of duplicated objects.
Instead of generated from dense grids, the kernels in \methodname~are a set of learnable parameters updated by their corresponding contents in the image.
\methodname~does not need to handle duplicated kernels because its kernels learn to focus on different regions of the image in training,
constrained by the bipartite matching strategy that builds a one-to-one mapping between the kernels and instances.
\vspace{-1mm}\section{Methodology}\label{sec:methods}\vspace{-1mm}
We consider various segmentation tasks through a unified perspective of kernels.
The proposed \shortname~uses a set of kernels to assign each pixel to either a potential instance or a semantic class (Sec.~\ref{sec:kernel_formulation}).
To enhance the discriminative capability of kernels, we contribute a way to update the static kernels by the contents in their partitioned pixel groups (Sec.~\ref{sec:kernel_head}).
We adopt the bipartite matching strategy to train instance kernels in an end-to-end manner (Sec.~\ref{sec:kernel_training}).
\shortname~can be applied seamlessly to semantic, instance, and panoptic segmentation as described in Sec.~\ref{sec:seg_archs}.

\vspace{-1mm}\subsection{K-Net}\label{sec:kernel_formulation}\vspace{-1mm}
Despite the different definitions of a `meaningful group', all segmentation tasks essentially assign each pixel to one of the predefined meaningful groups~\cite{szeliski2010computer}.
As the number of groups in an image is typically assumed finite, we can set the maximum group number of a segmentation task as $N$.
For example, there are $N$ pre-defined semantic classes for semantic segmentation or at most $N$ objects in an image for instance segmentation.
For panoptic segmentation, $N$ is the total number of stuff classes and objects in an image.
Therefore, we can use $N$ kernels to partition an image into $N$ groups,
where each kernel is responsible to find the pixels belonging to its corresponding group.
Specifically, given an input feature map $F \in R^{B\times C \times H \times W}$ of $B$ images, produced by a deep neural network,
we only need $N$ kernels $K \in R^{N \times C}$ to perform convolution with $F$ to obtain the corresponding segmentation prediction $M \in R^{B\times N \times H \times W}$ as
\begin{equation}
	M  = \sigma(K \ast F),
\end{equation}
where $C$, $H$, and $W$ are the number of channels, height, and width of the feature map, respectively.
The activation function $\sigma$ can be softmax function if we want to assign each pixel to only one of the kernels (usually used in semantic segmentation).
Sigmoid function can also be used as activation function if we allow one pixel belong to multiple masks, which results on $N$ binary masks by setting a threshold like 0.5 on the activation map (usually used in instance segmentation).

This formulation has already dominated semantic segmentation for years~\cite{FCN, PSPNet, deeplabv3}. In semantic segmentation, each kernel is responsible to find all pixels of a similar class across images.
Whereas in instance segmentation, each pixel group corresponds to an object.
However, previous methods separate instances by extra steps~\cite{mask_rcnn, associative_embedding, instance_cut} instead of by kernels.


This paper is the first study that explores if the notion of kernels in semantic segmentation is equally applicable to instance segmentation, and more generally panoptic segmentation. 
To separate instances by kernels, each kernel in \methodname~only segments at most one object in an image (Fig.~\ref{fig:teaser}-(b)).
In this way, \methodname~distinguishes instances and performs segmentation simultaneously, achieving instance segmentation in one pass without extra steps.
For simplicity, we call these kernels as \emph{semantic} and \emph{instance} kernels in this paper for semantic and instance segmentation, respectively.
A simple combination of instance kernels and semantic kernels can naturally preform panoptic segmentation that either assigns a pixel to an instance ID or a class of stuff (Fig.~\ref{fig:teaser}-(c)).

\begin{wrapfigure}{r}{0.55\textwidth}
	\vspace{-15pt}
	\centering
	\includegraphics[width=\linewidth]{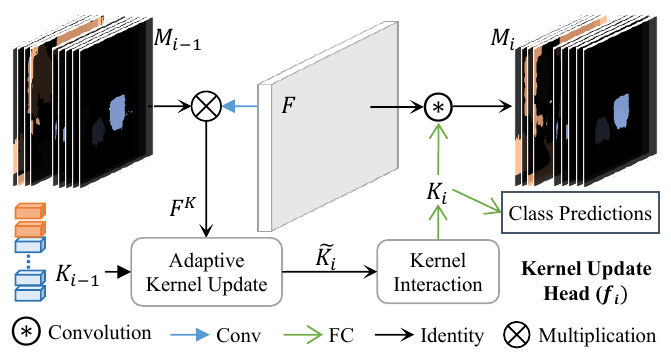}
	\vspace{-18pt}
	\caption{\small{
		Kernel Update Head.
		}}\label{fig:kernel_update_head}
	\vspace{-18pt}
\end{wrapfigure}

\vspace{-1mm}\subsection{Group-Aware Kernels}\label{sec:kernel_head}\vspace{-1mm}
Despite the simplicity of \methodname, separating instances directly by kernels is non-trivial.
Because instance kernels need to discriminate objects that vary in scale and appearance within and across images. Without a common and explicit characteristic like semantic categories,
the instance kernels need stronger discriminative ability than static kernels.

\begin{figure}[t]
	\centering\includegraphics[width=\linewidth]{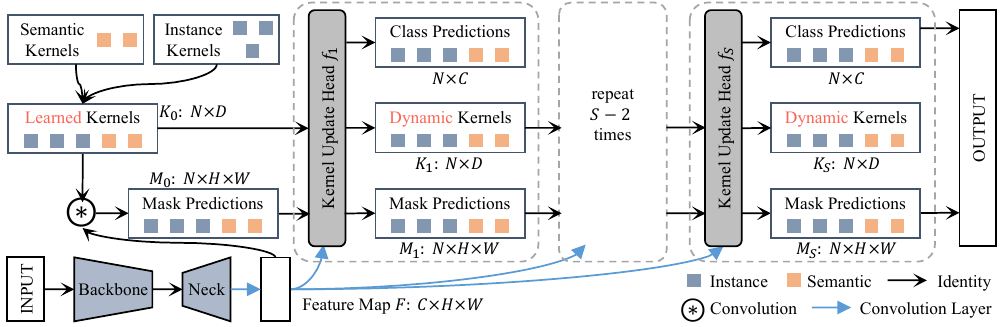}
	\vspace{-15pt}
	\caption{\small{
			\textbf{\methodname~for panoptic segmentaion.} A set of learned kernels first performs convolution with the feature map $F$ to  predict masks $M_0$.
			Then the kernel update head takes the mask predictions $M_0$, learned kernels $K_0$, and feature map $F$ as input and produce class predictions, group-aware (dynamic) kernels, and mask predictions.
			The produced mask prediction, dynamic kernels, and feature map $F$ are sent to the next kernel update head.
			This process is performed iteratively to progressively refine the kernels and the mask predictions.
		}}\label{fig:full_pipeline}
	\vspace{-9pt}
\end{figure}

To overcome this challenge, we contribute an approach to make the kernel conditional on their corresponding pixel groups, through a kernel update head, as shown in Fig.~\ref{fig:kernel_update_head}.
The kernel update head $f_{i}$ contains three key steps:
group feature assembling, adaptive kernel update, and kernel interaction.
Firstly, the group feature $F^K$ for each pixel group is assembled using the mask prediction $M_{i-1}$.
As it is the content of each individual groups that distinguishes them from each other,
$F^K$ is used to update their corresponding kernel $K_{i-1}$ adaptively.
After that, the kernel interacts with each other to comprehensively model the image context.
Finally, the obtained group-aware kernels $K_{i}$ perform convolution over feature map $F$ to obtain more accurate mask prediction $M_{i}$.
As shown in Fig.~\ref{fig:full_pipeline}, this process can be conducted iteratively because a finer partition usually reduces the noise in group features,
which results in more discriminative kernels. This process is formulated as
\begin{equation}
	K_{i}, M_{i} = f_{i}\left(M_{i-1}, K_{i-1}, F\right).
\end{equation}
Notably, the kernel update head with the iterative refinement is universal as it does not rely on the characteristic of kernels.
Thus, it can enhance not only instance kernels but also semantic kernels. We detail the three steps as follows.

\myparagraph{Group Feature Assembling.}
The kernel update head first assembles the features of each group, which will be adopted later to make the kernels group-aware.
As the mask of each kernel in $M_{i-1}$ essentially defines whether or not a pixel belongs to the kernel's related group,
we can assemble the feature $F^K$ for $K_{i-1}$ by multiplying the feature map $F$ with the $M_{i-1}$ as
\begin{equation}
	F^{K}  = \sum_{u}^{H}\sum_{v}^{W} M_{i-1}(u, v) \cdot F(u, v), F^{K} \in R^{B \times N \times C},
\end{equation}
where $B$ is the batch size, $N$ is the number of kernels, and $C$ is the number of channels.

\myparagraph{Adaptive Feature Update.}
The kernel update head then updates the kernels using the obtained $F^{K}$ to improve the representation ability of kernels.
As the mask $M_{i-1}$ may not be accurate, which is more common the case, the feature of each group may also contain noises introduced by pixels from other groups.
To reduce the adverse effect of the noise in group features, we devise an adaptive kernel update strategy.
Specifically, we first conduct element-wise multiplication between $F^{K}$ and $K_{i-1}$ as
\begin{equation}
		F^G = \phi_1(F^{K}) \otimes \phi_2(K_{i-1}), F^{G} \in R^{B \times N \times C},
\end{equation}
where $\phi_1$ and $\phi_2$ are linear transformations.
Then the head learns two gates, ${G}^{F}$ and ${G}^{K}$,
which adapt the contribution from $F^{K}$ and $K_{i-1}$ to the updated kernel $\tilde{K}$, respectively.
The formulation is
\begin{equation}
	\begin{aligned}
		{G}^{K} = \sigma(\psi_1(F^G)), {G}^{F} = \sigma(\psi_2(F^G)), \\
		\tilde{K} = {G}^{F} \otimes \psi_3(F^{K})  + {G}^{K} \otimes \psi_4(K_{i-1}),
	\end{aligned}
\end{equation}
where $\psi_{n}, n=1, ..., 4$ are different fully connected (FC) layers followed by LayerNorm (LN) and $\sigma$ is the Sigmoid function.
$\tilde{K}$ is then used in kernel interaction.

The gate learned here plays a role like the self-attention mechanism in Transformer~\cite{transformer},
whose output is computed as a weighted summation of the values.
In Transformer, the weight assigned to each value is usually computed by a compatibility function dot-product of the queries and keys.
Similarly, adaptive kernel update essentially performs weighted summation of kernel features $K_{i-1}$ and group features $F^G$.
Their weight ${G}^{K}$ and ${G}^{F}$ are computed by element-wise multiplication, which can be regarded as another kind of compatibility function.

\myparagraph{Kernel Interaction.}
Interaction among kernels is important to inform each kernel with contextual information from other groups.
Such information allows the kernel to implicitly model and exploit the relationship between groups of an image.
To this end, we add a kernel interaction process to obtain the new kernels $K_i$ given the updated kernels $\tilde{K}$.
Here we simply adopt Multi-Head Attention~\cite{transformer} followed by a Feed-Forward Neural Network, which has been proven effective in previous works~\cite{DETR, transformer}.
The output $K_i$ of kernel interaction is then used to generate a new mask prediction through $M_{i} = g_{i}\left(K_{i}\right) * F$,
where $g_{i}$ is an FC-LN-ReLU layer followed by an FC layer.
$K_{i}$ will also be used to predict classification scores in instance and panoptic segmentation.


\subsection{Training Instance Kernels}\label{sec:kernel_training}

While each semantic kernel can be assigned to a constant semantic class, there lacks an explicit rule to assign varying number of targets to instance kernels.
In this work, we adopt bipartite matching strategy and set prediction loss~\cite{set_prediction_loss, DETR} to train instance kernels in an end-to-end manner.
Different from previous works~\cite{set_prediction_loss, DETR} that rely on boxes, the learning of instance kernels is purely mask-driven 
because the inference of \methodname~is naturally box-free.

\myparagraph{Loss Functions.}
The loss function for instance kernels is written as
$L_{K} = \lambda_{cls}L_{cls} + \lambda_{ce}L_{ce} + \lambda_{dice}L_{dice}$,
where $L_{cls}$ is Focal loss~\cite{lin2017_focal} for classification,
and $L_{ce}$ and $L_{dice}$ are CrossEntropy (CE) loss and Dice loss~\cite{dice_loss} for segmentation, respectively.
Given that each instance only occupies a small region in an image,
CE loss is insufficient to handle the highly imbalanced learning targets of masks.
Therefore, we apply Dice loss~\cite{dice_loss} to handle this issue following previous works~\cite{solo, solov2, tian2020conditional}.


\myparagraph{Mask-based Hungarian Assignment.}
We adopt Hungarian assignment strategy used in~\cite{set_prediction_loss, DETR} for target assignment to train \methodname~in an end-to-end manner.
It builds a one-to-one mapping between the predicted instance masks and the ground-truth (GT) instances based on the matching costs.
The matching cost is calculated between the mask and GT pairs in a similar manner as the training loss.

\subsection{Applications to Various Segmentation Tasks}\label{sec:seg_archs}
\myparagraph{Panoptic Segmentation.}
For panoptic segmentation, the kernels are composed of instance kernels $K_{0}^{ins}$ and semantic kernels $K_{0}^{sem}$ as shown in Fig.~\ref{fig:full_pipeline}.
We adopt semantic FPN~\cite{panoptic_fpn} for producing high resolution feature map $F$,
except that we add positional encoding used in~\cite{transformer, DETR, DeformableDETR} to enhance the positional information.
Specifically, given the feature maps $P2, P3, P4, P5$ produced by FPN~\cite{lin2017_fpn}, positional encoding is computed based on the feature map size of $P5$, and it is added with $P5$.
Then semantic FPN~\cite{panoptic_fpn} is used to produce the final feature map.

As semantic segmentation mainly relies on semantic information for per-pixel classification, while instance segmentation prefers accurate localization information to separate instances,
we use two separate branches to generate the features $F^{ins}$ and $F^{sem}$ to perform convolution with $K_{0}^{ins}$ and $K_{0}^{sem}$ for generating instance and semantic masks $M_0^{ins}$ and $M_0^{sem}$, respectively.
Notably, it is unnecessary to produce `thing' and `stuff' masks initially from different branches to produce a reasonable performance.
Such a design is consistent with previous practices~\cite{panoptic_fpn, solov2} and empirically yields better performance (about 1\% PQ).

We then construct $M_0$, $K_0$, and $F$ as the inputs of kernel update head to dynamically update the kernels and refine the panoptic mask prediction.
Because `things' are already separated by instance masks in $M_0^{ins}$, while $M_0^{sem}$ contains the semantic masks of both `things' and `stuff',
we select $M_0^{st}$, the masks of stuff categories from $M_0^{sem}$,
and directly concatenate it with $M_0^{ins}$ to form the panoptic mask prediction $M_0$.
Due to similar reason, we only select and concatenate the kernels of stuff classes in $K_0^{sem}$ with $K_0^{ins}$ to form the panoptic kernels $K_0$.
To exploit the complementary semantic information in $F^{sem}$ and localization information in $F^{ins}$,
we add them together to obtain $F$ as the input feature map of the kernel update head.
With $M_0$, $K_0$, and $F$, the kernel update head $f_1$ can produce group-aware kernels $K_1$ and mask $M_1$.
Then kernels and masks are iteratively by $S$ times and finally we can obtain the mask prediction $M_S$.

To produce the final panoptic segmentation results, we paste thing and stuff masks in a mixed order following MaskFormer~\cite{cheng2021maskformer}.
We also find it necessary in K-Net to firstly sort the pasting order of masks based on their classification scores for further filtering out lower-confident mask predictions.
Such a method empirically performs better (about 1\% PQ) than the previous strategy that pasting thing and stuff masks separately~\cite{panoptic_fpn, panoptic_fcn}.

\myparagraph{Instance Segmentation.}
In the similar framework, we simply remove the concatenation process of kernels and masks to perform instance segmentation.
We did not remove the semantic segmentation branch as the semantic information is still complementary for instance segmentation.
Note that in this case, the semantic segmentation branch does not use extra annotations.
The ground truth of semantic segmentation is built by converting instance masks to their corresponding class labels.

\myparagraph{Semantic Segmentation.}
As \methodname~does not rely on specific architectures of model representation,
\methodname~can perform semantic segmentation by simply appending its kernel update head to any existing semantic segmentation methods~\cite{PSPNet, FCN, deeplabv3, upernet} that rely on semantic kernels.


\vspace{-1mm}\section{Experiments}\label{sec:Experiments}\vspace{-1mm}


\myparagraph{Dataset and Metrics.}
For panoptic and instance segmentation, we perform experiments on the challenging COCO dataset~\cite{lin2014coco}.
All models are trained on the \texttt{train2017} split and evaluated on the \texttt{val2017} split.
The panoptic segmentation results are evaluated by the PQ metric~\cite{panoptic}.
We also report the performance of thing and stuff, noted as PQ\things, PQ\stuff, respectively, for thorough evaluation.
The instance segmentation results are evaluated by mask AP~\cite{lin2014coco}.
The AP for small, medium and large objects are noted as AP$_{s}$, AP$_{m}$, and AP$_{l}$, respectively.
The AP at mask IoU thresholds 0.5 and 0.75 are also reported as AP$_{50}$ and AP$_{75}$, respectively.
For semantic segmentation, we conduct experiments on the challenging ADE20K dataset~\cite{ADE20K} and report mIoU to evaluate the segmentation quality.
All models are trained on the \texttt{train} split and evaluated on the \texttt{validation} split.

\myparagraph{Implementation Details.}
For panoptic and instance segmentation, we implement \methodname~with MMDetection~\cite{mmdetection}.
In the ablation study, the model is trained with a batch size of 16 for 12 epochs.
The learning rate is 0.0001, and it is decreased by 0.1 after 8 and 11 epochs, respectively. 
We use AdamW~\cite{ADAMW} with a weight decay of 0.05.
For data augmentation in training, we adopt horizontal flip augmentation with a single scale.
The long edge and short edge of images are resized to 1333 and 800, respectively, without changing the aspect ratio.
When comparing with other frameworks,
we use multi-scale training with a longer schedule (36 epochs) for fair comparisons~\cite{mmdetection}.
The short edge of images is randomly sampled from $[640, 800]$~\cite{He_2019_ICCV}.

For semantic segmentation, we implement \methodname~with MMSegmentation~\cite{mmseg2020} and train it with 80,000 iterations.
As AdamW~\cite{ADAMW} empirically works better than SGD, we use AdamW with a weight decay of 0.0005 by default on both the baselines and \methodname~for a fair comparison.
The initial learning rate is 0.0001, and it is decayed by 0.1 after 60000 and 72000 iterations, respectively.
More details are provided in the appendix.

\myparagraph{Model Hyperparameters.}
In the ablation study, we adopt ResNet-50~\cite{He_2016} backbone with FPN~\cite{lin2017_fpn}.
For panoptic and instance segmentation, we use $\lambda_{cls} = 2$ for Focal loss following previous methods~\cite{DeformableDETR}, and empirically find $\lambda_{seg}=1$, $\lambda_{ce} = 1$, $\lambda_{dice} = 4$ work best.
For efficiency, the default number of instance kernels is 100.
For semantic segmentation, $N$ equals to the number of classes of the dataset, which is 150 in ADE20K and 133 in COCO dataset.
The number of rounds of iterative kernel update is set to three by default for all segmentation tasks.

\begin{table*}[t]
	\caption{\small{
    Comparisons with state-of-the-art panoptic segmentation methods on COCO dataset}
	}\label{tab:benchmark:panoptic}
  \centering
  \footnotesize
	\vspace{-6pt}
  \scalebox{0.9}{\tablestyle{2pt}{1.0}{
    \begin{tabular}{c|c|c|c|c|ccc}
      Framework & Backbone & Box-free & NMS-free & Epochs & PQ & PQ\things & PQ\stuff \\
      \shline
      \multicolumn{8}{c}{\texttt{val}} \\
      \hline
      Panoptic-DeepLab~\cite{panoptic_deeplab} &Xception-71&&&$\sim$1000&39.7&43.9&33.2 \\
      Panoptic FPN~\cite{panoptic_fpn} & R50-FPN &&&36&41.5 & 48.5& 31.1 \\
      SOLOv2~\cite{solov2} & R50-FPN &\checkmark && 36 & 42.1 &49.6 &30.7 \\
      DETR~\cite{DETR}$^{\dagger}$ & R50& &\checkmark& 300 + 25 &43.4 & 48.2& 36.3 \\
      Unifying~\cite{unifying_CVPR2020}& R50-FPN &&& $\sim$27 & 43.4 & 48.6 & 35.5 \\
      Panoptic FCN~\cite{panoptic_fcn} & R50-FPN &\checkmark  &\checkmark & 36 & 43.6& 49.3& 35.0 \\
      \shortname& R50-FPN &\checkmark &\checkmark  & 36 & \bd{47.1}&\bd{51.7}& \bd{40.3} \\ \hline
      \multirow{3}{*}{\shortname}& R101-FPN &\checkmark&\checkmark& 36 &49.6&55.1&41.4 \\
      & R101-FPN-DCN &\checkmark &\checkmark &36 &48.3&54.0&39.7 \\
      & Swin-L~\cite{SwinTransformer} &\checkmark &\checkmark & 36&54.6&60.2&46.0 \\
      \hline
      \multicolumn{8}{c}{\texttt{test-dev}} \\ \hline
      Panoptic-DeepLab&Xception-71&\checkmark&&$\sim$1000&41.4&45.1&35.9 \\
      Panoptic FPN & R101-FPN &&&36&43.5&50.8&32.5 \\
      Panoptic FCN & R101-FPN &\checkmark &\checkmark & 36 &45.5&51.4&36.4 \\
      DETR & R101 &&\checkmark& 300 + 25 &46.0&-&- \\
      UPSNet~\cite{UPSNet}&R101-FPN-DCN &&&36&46.6&53.2&36.7 \\
      Unifying~\cite{unifying_CVPR2020}& R101-FPN-DCN &&& $\sim$27 &47.2&53.5&37.7 \\
      \shortname& R101-FPN &\checkmark &\checkmark  & 36 &47.0&52.8&38.2 \\
      \shortname& R101-FPN-DCN &\checkmark &\checkmark  & 36 &\bd{48.3}&\bd{54.0}&\bd{39.7} \\ \hline
      MaX-DeepLab-L~\cite{max_deeplab} & Max-L &\checkmark &\checkmark & 54 &51.3 &57.2 &42.4 \\
      MaskFormer~\cite{cheng2021maskformer} & Swin-L~\cite{SwinTransformer} &\checkmark &\checkmark & 300 &53.3 &59.1 &44.5 \\
      Panoptic SegfFormer~\cite{panoptic_segformer} &PVTv2-B5~\cite{PVTv2}&\checkmark &\checkmark&50&54.4&61.1&44.3 \\
      \shortname& Swin-L &\checkmark &\checkmark & 36 &\bd{55.2}&\bd{61.2}&\bd{46.2} \\
		\end{tabular}}}
	\vspace{-18pt}
\end{table*}

\subsection{Benchmark Results}
\myparagraph{Panoptic Segmentation.}
We first benchmark \methodname~with other panoptic segmentation frameworks in Table~\ref{tab:benchmark:panoptic}.
\methodname~surpasses the previous state-of-the-art box-based method~\cite{unifying_CVPR2020} and box/NMS-free method~\cite{panoptic_fcn} by 1.7 and 1.5 PQ on \texttt{val} split, respectively.
On the \texttt{test-dev} split, \methodname~with ResNet-101-FPN backbone even obtains better results than that of UPSNet~\cite{UPSNet}, which uses Deformable Convolution Network (DCN)~\cite{dcn}.
\methodname~equipped with DCN surpasses the previous method~\cite{unifying_CVPR2020} by 1.1 PQ.
Without bells and whistles, \methodname~obtains new state-of-the-art single-model performance with Swin Transformer~\cite{SwinTransformer} serving as the backbone.

We also compare \methodname~with concurrent work Max-DeepLab~\cite{max_deeplab}, MaskFormer~\cite{cheng2021maskformer}, and Panoptic SegFormer~\cite{panoptic_segformer}.
\methodname~surpasses these methods with the least training epochs (36), taking only about 44 GPU days (roughly 2 days and 18 hours with 16 GPUs).
Note that only 100 instance kernels and Swin Transformer with window size 7 are used here for efficiency.
\methodname~could obtain a higher performance with more instance kernels (Sec. \ref{sec:exp:ablation}), Swin Transformer with window size 12 (used in MaskFormer~\cite{cheng2021maskformer}),
as well as an extended training schedule with aggressive data augmentation used in previous work~\cite{DETR}.

\begin{table*}[t]
	\caption{
    \small{Comparisons with state-of-the-art instance segmentation methods on COCO dataset.
    `P. (M)' indicates the number of parameters in the model, and the counting unit is million}
	}\label{tab:benchmark:instance}
  \centering
  \footnotesize
	\vspace{-6pt}
  \scalebox{0.88}{\tablestyle{2pt}{1.0}{
    \begin{tabular}{c|c|c|c|c|c|cc|ccc|c|c}
      Method & Backbone & Box-free & NMS-free & Epochs &AP$\uparrow$&AP$_{50}$ & AP$_{70}$& AP$_{s}$ & AP$_{m}$ & AP$_{l}$ & FPS$\uparrow$& P. (M) $\downarrow$ \\ \shline
      \multicolumn{13}{c}{\texttt{val2017}} \\\hline
      SOLO~\cite{solo}& R-50-FPN &\checkmark && 36 &35.8 &56.7 & 37.9 & 14.3 & 39.3 & 53.2 & 12.7&36.08 \\
      Mask R-CNN~\cite{mask_rcnn} & R-50-FPN & & & 36 &37.1& 58.5& 39.7&18.7 & 39.6 & 53.9 &17.5&44.17 \\
      SOLOv2~\cite{solov2}& R-50-FPN &\checkmark && 36&37.5 &58.2&40.0&15.8&41.4&56.6&17.7&\bd{33.89} \\
      CondInst~\cite{tian2020conditional}& R-50-FPN & && 36 &37.5&58.5&40.1&18.7&41.0&53.3&14.0&46.37 \\
      Cascade Mask R-CNN~\cite{cascade_rcnn}& R-50-FPN&&&36&38.5&59.7&\bd{41.8}&\bd{19.3}& 41.1 & 55.6 &10.3&77.10 \\
      \shortname& R-50-FPN &\checkmark &\checkmark & 36&37.8& 60.3& 39.9& 16.9& 41.2& 57.5 & \bd{21.2} &37.26\\
      \shortname-N256& R-50-FPN &\checkmark &\checkmark & 36&\bd{38.6}&\bd{60.9}& 41.0& 19.1&\bd{42.0}&\bd{57.7} & 19.8&37.30\\
      \hline

      \multicolumn{13}{c}{\texttt{test-dev}} \\\hline
      SOLO& R-50-FPN &\checkmark && 72 &36.8&58.6&39.0&15.9&39.5&52.1&12.7&36.08 \\
      Mask R-CNN & R-50-FPN & & &36&37.4&59.5&40.1&18.6&39.8&51.6&17.5&44.17 \\
      CondInst &R-50-FPN & & &36&37.8&59.2&40.4&18.2&40.3&52.7 &14.0&46.37 \\
      SOLOv2& R-50-FPN &\checkmark &&  36&38.2&59.3&40.9&16.0&41.2&55.4&17.7&\bd{33.89} \\
      Cascade Mask R-CNN& R-50-FPN&&&36&38.8&60.4&\bd{42.0}&\bd{19.4}&40.9&53.9&10.3&77.10 \\
      \shortname& R-50-FPN &\checkmark &\checkmark & 36&38.4&61.2&40.9&17.4&40.7&56.2  & \bd{21.2}&37.26\\
      \shortname-N256& R-50-FPN &\checkmark &\checkmark & 36&\bd{39.1}&\bd{61.7}&41.8&18.2&\bd{41.4}&\bd{56.6} & 19.8&37.30\\
      \hline
      SOLO& R-101-FPN &\checkmark &&  72 &37.8&59.5&40.4&16.4&40.6&54.2&10.7&55.07 \\
      Mask R-CNN & R-101-FPN & & &36&38.8&60.8&41.8&19.1&41.2&54.3 & 14.3&63.16 \\
      CondInst &R-101-FPN & & &36&38.9&60.6&41.8&18.8&41.8&54.4&11.0&\bd{52.83} \\
      SOLOv2& R-101-FPN &\checkmark &&  36&39.5&60.8&42.6&16.7&43.0&57.4 & 14.3&65.36 \\
      Cascade Mask R-CNN& R-101-FPN&&&36&39.9&61.6&43.3&\bd{19.8}&42.1&55.7 & 9.5&96.09 \\
      \shortname &R-101-FPN &\checkmark &\checkmark & 36&40.1&62.8&43.1&18.7&42.7&58.8 & \bd{16.2}&56.25 \\
      \shortname-N256&R-101-FPN &\checkmark &\checkmark & 36&\bd{40.6}&\bd{63.3}&\bd{43.7}&18.8&\bd{43.3}&\bd{59.0} &15.5&56.29 \\
		\end{tabular}}}
	\vspace{-12pt}
\end{table*}

\begin{table*}[t]
  \caption{\small{Results of \methodname~on ADE20K semantic segmentation dataset}} \label{tab:benchmark:semantic}
  \vspace{-6pt}
  \centering
  \footnotesize
  \begin{minipage}[b]{.44\linewidth}
    \subcaption{\small{Improvements of \methodname~on different architectures}}\label{tab:benchmark:semantic:val}
    \vspace{-3pt}
    \centering
    \scalebox{0.9}{\tablestyle{2pt}{1.0}\begin{tabular}{l|c|c}
      Method & Backbone & Val mIoU  \\
      \shline
      FCN~\cite{FCN}& R50 & 36.7 \\
      FCN + \shortname& R50 & 43.3 \dt{(+6.6)}  \\\hline
      PSPNet~\cite{PSPNet} & R50 & 42.6 \\
      PSPNet + \shortname & R50 & 43.9 \dt{(+1.3)} \\\hline
      DLab.v3~\cite{deeplabv3} & R50 & 43.5 \\
      DLab.v3 + \shortname & R50 & 44.6 \dt{(+1.1)} \\\hline
      UperNet~\cite{upernet} & R50 & 42.4 \\
      UperNet + \shortname & R50 & 43.6 \dt{(+1.2)}  \\\hline
      UperNet & Swin-L & 50.6 \\
      UperNet + \shortname & Swin-L & 52 \dt{(+1.4)} \\
      \end{tabular}}
  \end{minipage}\hspace{3mm}
  \begin{minipage}[b]{.48\linewidth}
    \subcaption{\small{Comparisons with state-of-the-art methods. Results marked by $^{\dagger}$ use larger image sizes}}\label{tab:benchmark:semantic:test}
    \vspace{-3pt}
    \centering
    \scalebox{0.9}{\tablestyle{2pt}{1.0}\begin{tabular}{l|c|c}
      Method & Backbone & Val mIoU \\
      \shline
      OCRNet~\cite{OCRNet} & HRNet-W48 & 44.9 \\
      PSPNet~\cite{PSPNet} & R101 & 45.4 \\
      PSANet~\cite{psanet} & R101 & 45.4 \\
      DNL~\cite{DNL} & R101 & 45.8 \\
      DLab.v3~\cite{deeplabv3} & R101 & 46.7 \\
      DLab.v3+\cite{deeplabv3plus} & S-101~\cite{resnest} & 47.3 \\
      SETR~\cite{SETR} & ViT-L~\cite{VIT} & 48.6 \\
      UperNet$^{\dagger}$ & Swin-L & 53.5 \\\hline
      UperNet + \shortname & Swin-L & 53.3\\
      UperNet + \shortname$^{\dagger}$ & Swin-L & \bd{54.3}\\
      \end{tabular}}
  \end{minipage}
\vspace{-18pt}
\end{table*}

\myparagraph{Instance Segmentation.}
We compare \methodname~with other instance segmentation frameworks~\cite{mask_rcnn, tian2020conditional, chen2019tensormask} in Table~\ref{tab:benchmark:instance}.
More details are provided in the appendix.
As the only box-free and NMS-free method, \methodname~achieves better performance and faster inference speed than Mask R-CNN~\cite{mask_rcnn}, SOLO~\cite{solo}, SOLOv2~\cite{solov2} and CondInst~\cite{tian2020conditional},
indicated by the higher AP and frames per second (FPS).
We adopt 256 instance kernels (\methodname-N256 in the table) to compare with Cascade Mask R-CNN~\cite{cascade_rcnn}.
The performance of \methodname-N256 is on par with Cascade Mask R-CNN~\cite{cascade_rcnn} but enjoys a \bd{92.2\%} faster inference speed (19.8 v.s 10.3).

On COCO \texttt{test-dev} split, \methodname~with ResNet-101-FPN backbone obtains performance that is 0.9 AP better than Mask R-CNN~\cite{mask_rcnn}.
It also surpasses previous kernel-based approach CondInst~\cite{tian2020conditional} and SOLOv2~\cite{solov2} by 1.2 AP and 0.6 AP, respectively. 
With ResNet-101-FPN backbone, \methodname~surpasses Cascade Mask R-CNN with 100 and 256 instance kernels in both accuracy and speed by 0.2 AP and 6.7 FPS, and 0.7 AP and 6 FPS, respectively.

We also compare the number of parameters of these models in Table~\ref{tab:benchmark:instance}.
Though \methodname~does not have the least number of parameters, it is more lightweight than Cascade Mask R-CNN by approximately half number of the parameters (37.3 M vs. 77.1 M).

\myparagraph{Semantic Segmentation.} We apply \methodname~to existing frameworks~\cite{FCN, PSPNet, deeplabv3, upernet} that rely on static semantic kernels in Table~\ref{tab:benchmark:semantic:val}.
\methodname~consistently improves different frameworks.
Notably, \methodname~significantly improves FCN (\bd{6.6} mIoU).
This combination surpasses PSPNet and UperNet by 0.7 and 0.9 mIoU, respectively, and achieves performance comparable with DeepLab v3.
Furthermore, the effectiveness of \methodname~does not saturate with strong model representation, 
as it still brings significant improvement (1.4 mIoU) over UperNet with Swin Transformer~\cite{SwinTransformer}.
The results suggest the versatility and effectiveness of \methodname~for semantic segmentation.

In Table~\ref{tab:benchmark:semantic:test}, we further compare \methodname~with other state-of-the-art methods~\cite{SETR, deeplabv3plus} with test-time augmentation on the validation set.
With the input of 512$\times$512, \methodname~already achieves state-of-the-art performance.
With a larger input of 640$\times$640 following previous method~\cite{SwinTransformer} during training and testing,
\methodname~with UperNet and Swin Transformer achieves new state-of-the-art single model performance, which is 0.8 mIoU higher than the previous one.

\begin{table*}[t]
  \caption{\small{Ablation studies of \methodname~on instance segmentation}}
  \label{table:ablation}
  \vspace{-6pt}
  \centering
  \footnotesize
  \begin{minipage}[b]{.48\linewidth}
    \subcaption{\small{Adaptive Kernel Update (A. K. U.) and Kernel Interaction (K. I.)}}\label{tab:ablation:head}
    \vspace{-3pt}
    \centering
    \scalebox{0.88}{\tablestyle{8pt}{1.0}
      \begin{tabular}{c|c|ccc}
        A. K. U. & K. I. &AP& AP$_{50}$ & AP$_{75}$ \\
        \shline
        &&10.0&18.2&9.6 \\
        \checkmark &&22.6&37.3&23.5 \\
        & \checkmark &31.2&52.0&32.4 \\
        \checkmark & \checkmark &34.1&55.3&35.7 \\
      \end{tabular}}
  \end{minipage}\hspace{.02\linewidth}
  \begin{minipage}[b]{.48\linewidth}
    \subcaption{\small{Positional Encoding (P. E.) and Coordinate Convolution (Coors.)}}\label{tab:ablation:positional_encoding}
    \vspace{-3pt}
    \centering
    \scalebox{0.88}{\tablestyle{8pt}{1.0}\begin{tabular}{c|c|ccc}
      Coors. & P. E. &AP& AP$_{50}$ & AP$_{75}$ \\
      \shline
      &&30.9&51.7&31.6\\
      \checkmark &&34.0&55.4&35.6 \\
      & \checkmark &34.1&55.3&35.7 \\
      \checkmark & \checkmark &34.0&	55.1&35.8\\
      \end{tabular}}
  \end{minipage}
  \vspace{12pt}
  \begin{minipage}[b]{.48\linewidth}
    \subcaption{\small{Number of rounds of kernel update}}\label{tab:ablation:stage_num}
    \vspace{-3pt}
    \centering
    \scalebox{0.88}{\tablestyle{6.5pt}{1.0}\begin{tabular}{c|ccc|c}
      Stage Number &AP& AP$_{50}$ & AP$_{75}$ & FPS \\
      \shline
      1 &21.8	&37.3	&22.1 &24.0 \\
      2 &32.1	&52.3	&33.5 &22.7 \\
      3 &34.1 &55.3 &35.7 &21.2 \\
      4 &34.5	&56.5	&35.7 &20.1 \\
      5 &34.5	&56.5	&35.9 &18.9 \\
      \end{tabular}}
  \end{minipage}\hspace{.02\linewidth}
  \begin{minipage}[b]{.48\linewidth}
    \subcaption{\small{Numbers of instance kernels}}\label{tab:ablation:kernel_num}
    \vspace{-3pt}
    \centering
    \scalebox{0.88}{\tablestyle{9pt}{1.0}\begin{tabular}{c|ccc|c}
      $N$ &AP& AP$_{50}$ & AP$_{75}$ & FPS \\
      \shline
      50 &32.7	&53.7	&34.1 &21.6 \\
      64 &33.6	&54.8	&35.1 &21.6 \\
      100&34.1  &55.3 &35.7 &21.2 \\
      128&34.3  &55.6 &35.8 &20.7 \\
      256&34.7  &56.1 &36.3 &19.8 \\
      \end{tabular}}
  \end{minipage}
  \vspace{-24pt}
\end{table*}

\subsection{Ablation Study on Instance Segmentation}\label{sec:exp:ablation}
We conduct an ablation study on COCO instance segmentation dataset to evaluate the effectiveness of \methodname~in discriminating instances.
The conclusion is also applicable to other segmentation tasks since the design of \methodname~is universal to all segmentation tasks.

\myparagraph{Head Architecture.} We verify the components in the kernel update head in Table~\ref{tab:ablation:head}.
The results of without A. K. U. is obtained by updating kernels purely by $\tilde{K} = F_K + K_{i-1}$ followed by an FC-LN-ReLU layer.
The results indicates that both adaptive kernel update and kernel interaction
are necessary for high performance.

\myparagraph{Positional Information.} 
We study the necessity of positional information in Table ~\ref{tab:ablation:positional_encoding}.
The results show that positional information is beneficial, and positional encoding~\cite{transformer, DETR} works slightly better than coordinate convolution.
The combination of the two components does not bring additional improvements. The results justify the use of just positional encoding in our framework.

\myparagraph{Number of Stages.} We compare different kernel update rounds in Table~\ref{tab:ablation:stage_num}.
The results show that FPS decreases as the update rounds grow while the performance saturates beyond three stages.
\begin{wraptable}{r}{0.55\textwidth}
  \centering
  \vspace{-9pt}
  \caption{\small{Numbers of semantic kernels}}\label{tab:ablation:kernel_num:sem}
  \vspace{-3pt}
  \scalebox{0.88}{
    \tablestyle{3pt}{1.0}\begin{tabular}{c|c|c|c|c|c|c|c|c}
    Stage Number & 0&1&2&3&4&5&6&7 \\\shline
    mIoU &36.7&42.7&43.0&43.3&43.8&44.1&43.1&42.6
  \end{tabular}}
  \vspace{-9pt}
\end{wraptable}
Such a conclusion also holds for semantic segmentation as shown in Table \ref{tab:ablation:kernel_num:sem}. The performance of FCN $+$ K-Net on ADE20K dataset 
gradually increases as the increase of iteration number but also saturates after four iterations.

\myparagraph{Number of Kernels.} 
We further study the number of kernels in \methodname.
The results in Table~\ref{tab:ablation:kernel_num} reveal that 100 kernels are sufficient to achieve good performance.
The observation is expected for COCO dataset because most of the images in the dataset do not contain many objects (7.7 objects per image in average~\cite{lin2014coco}).
\methodname~consistently achieves better performance given more instance kernels since they improve the models' capacity in coping with complicated images.
However, a larger $N$ may lead to small performance gains and then get saturated (when $N=300, 512, 768$, we all get 34.9\% mAP).
Therefore, we select $N=100$ in other experiments for efficiency if without further specification.


\subsection{Visual Analysis}\label{sec:Analysis}

\begin{figure}[t]
    \centering
    \begin{minipage}[b]{.393\linewidth}
        \centering\includegraphics[width=\linewidth]{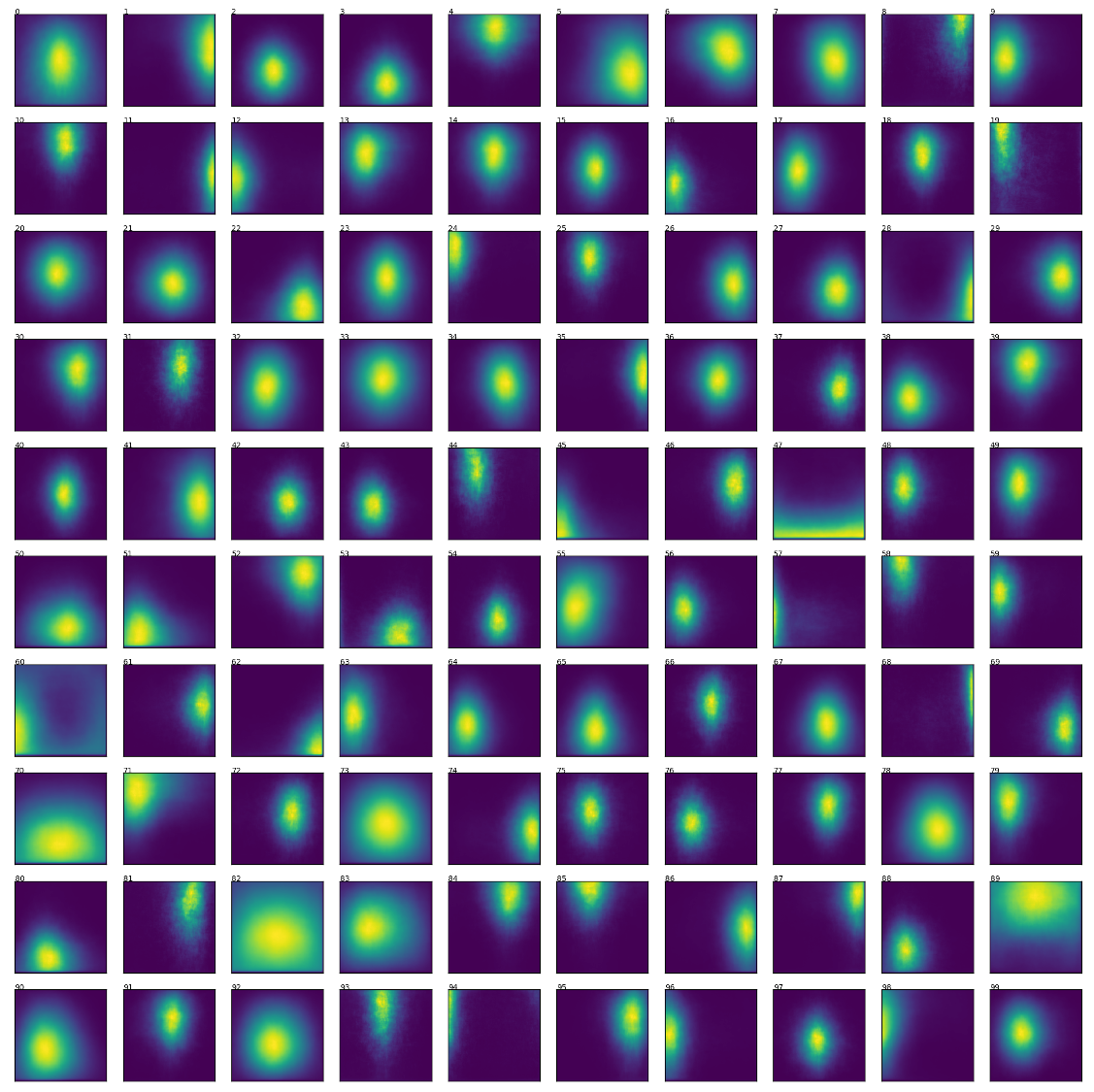}
	    \vspace{-12pt}
	    \subcaption{\footnotesize{
	    		Average activation over 5000 images.
	    	}}\label{fig:kernel_spatial_distribution}
    \end{minipage}\hspace{1mm}
    \begin{minipage}[b]{.58\linewidth}
        \centering\includegraphics[width=\linewidth]{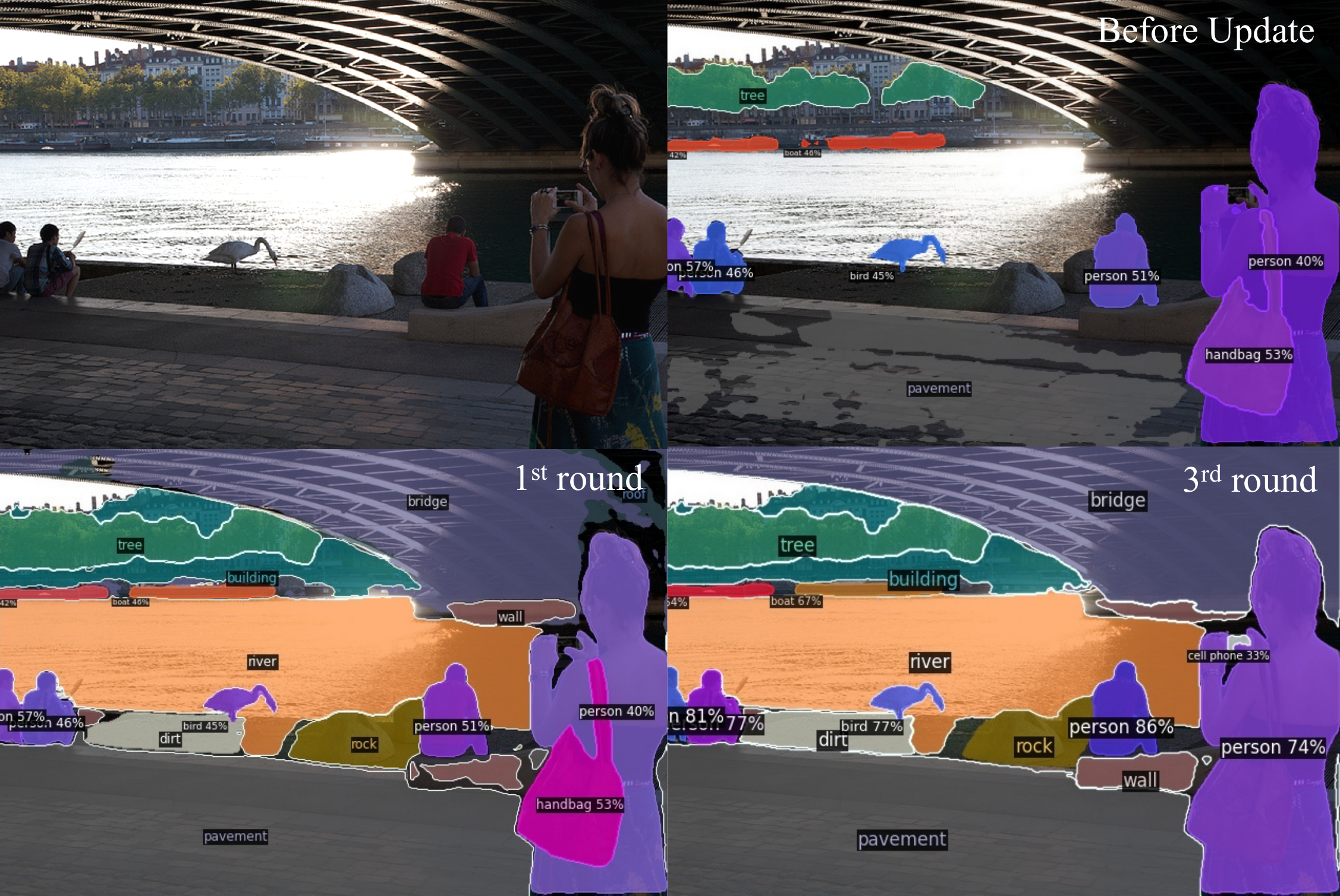}
	    \vspace{-12pt}
	    \subcaption{\footnotesize{
	    		Mask prediction before and after kernel update.
	    	}}\label{fig:kernel_mask_preds}
    \end{minipage}
    \vspace{-6pt}
    \caption{\small{
		Visual analysis of kernels and their masks. Best viewed in color and by zooming in.
	}}\label{fig:kernel_distribution}
	\vspace{-18pt}
\end{figure}

\myparagraph{Overall Distribution of Kernels.}
We carefully analyze the properties of instance kernels learned in \methodname~by analyzing the average of mask activations of the 100 instance kernels over the 5000 images in the \texttt{val} split.
All the masks are resized to have a similar resolution of 200 $\times$ 200 for the analysis.
As shown in Fig.~\ref{fig:kernel_spatial_distribution}, the learned kernels are meaningful. Different kernels specialize on different regions of the image and objects with different sizes,
while each kernel attends to objects of similar sizes at close locations across images.

\myparagraph{Masks Refined through Kernel Update.}
We further analyze how the mask predictions of kernels are refined through the kernel update in Fig.~\ref{fig:kernel_mask_preds}.
Here we take \methodname~for panoptic segmentation to thoroughly analyze both semantic and instance masks.
The masks produced by static kernels are incomplete, \eg, the masks of river and building are missed.
After kernel update, the contents are thoroughly covered by the segmentation masks, though the boundaries of masks are still unsatisfactory.
The boundaries are refined after more kernel update.
The classification confidences of instances also increase after kernel update.
More results are given in the appendix.

\vspace{-1mm}\section{Conclusion}\vspace{-1mm}

This paper explores instance kernels that can learn to separate instances during segmentation.
Thus, extra components that previously assist instance segmentation can be replaced by instance kernels, including bounding boxes, embedding generation, and hand-crafted post-processing like NMS, kernel fusion, and pixel grouping.
Such an attempt, for the first time, allows different image segmentation tasks to be tackled through a unified framework.
The framework, dubbed as \methodname, first partitions an image into different groups by learned static kernels, 
then iteratively refines these kernels and their partition of the image by the features assembled from their partitioned groups.
\methodname~obtains new state-of-the-art single-model performance on panoptic and semantic segmentation benchmarks
and surpasses the well-developed Cascade Mask R-CNN with the fastest inference speed among the recent instance segmentation frameworks.
We wish \methodname~and the analysis to pave the way for future research on unified image segmentation frameworks.

\appendix
\setcounter{table}{0} 
\setcounter{figure}{0}
\setcounter{equation}{0}
\renewcommand{\thetable}{A\arabic{table}}
\renewcommand\thefigure{A\arabic{figure}} 
\renewcommand\theequation{A\arabic{equation}}

\myparagraph{Acknowledgements.} 
This study is supported under the RIE2020 Industry Alignment Fund Industry Collaboration Projects (IAF-ICP) Funding Initiative, as well as cash and in-kind contribution from the industry partner(s).
It is also partially supported by the NTU NAP grant.
Jiangmiao Pang and Kai Chen are also supported by the Shanghai Committee of Science and Technology, China (Grant No. 20DZ1100800).
The authors would like to thank the valuable suggestions and comments by Jiaqi Wang, Rui Xu, and Xingxing Zou.
\newpage
\appendix
\begin{center}{\bf \Large Appendix}\end{center}\vspace{-1mm}

We first provide more implementation details for the three segmentation tasks of K-Net (Appendix~\ref{sec:appendix:implementation}).
Then we provide more benchmark details and discussion of comparison between K-Net and other methods (Appendix~\ref{sec:appendix:benchmark}).
We further analyze K-Net about its results of kernel update and failure cases (Appendix~\ref{sec:appendix:analysis}).
Last but not the least, we discuss the broader impact of~\methodname~(Appendix~\ref{sec:appendix:impact}).

\vspace{-1mm}\section{Implementation Details}\label{sec:appendix:implementation}\vspace{-1mm}
\myparagraph{Training details for semantic segmentation.}
We implement \methodname~based on MMSegmentation~\cite{mmseg2020} for experiments on semantic segmentation.
We use AdamW~\cite{ADAMW} with a weight decay of 0.0005 and train the model by 80000 iterations by default.
The initial learning rate is 0.0001, and it is decayed by 0.1 after 60000 and 72000 iterations, respectively.
This is different from the default training setting in MMSegmentation~\cite{mmseg2020} that uses SGD with momentum by 160000 iterations.
But our setting obtains similar performance as the default one.
Therefore, we apply AdamW with 80000 iterations to all the models in Table 3a of the main text for efficiency while keeping fair comparisons.
For data augmentation, we follow the default settings in MMSegmentation~\cite{mmseg2020}.
The long edge and short edge of images are resized to 2048 and 512, respectively, without changing the aspect ratio (described as 512 $\times$ 512 in the main text for short).
Then random crop, horizontal flip, and photometric distortion augmentations are adopted.

\vspace{-1mm}\section{Benchmark Results}\label{sec:appendix:benchmark}\vspace{-1mm}
\vspace{-1mm}\subsection{Instance Segmentation}\vspace{-1mm}
\myparagraph{Accuracy comparison.}
In Table 2 of the main text, we compare both accuracy and inference speed of \methodname~with previous methods.
For fair comparison, we re-implement Mask R-CNN~\cite{mask_rcnn} and Cascade Mask R-CNN~\cite{cascade_rcnn} with the multi-scale 3$\times$ training schedule~\cite{mmdetection, wu2019detectron2},
and submit their predictions to the evaluation server\footnote{\url{https://competitions.codalab.org/competitions/20796}} for obtaining their accuracies on the \texttt{test-dev} split.
For SOLO~\cite{solo}, SOLOv2~\cite{solov2}, and CondInst~\cite{tian2020conditional}, we test and report the accuracies of the models released in their official implementation~\cite{tian2019adelaidet}, which are trained by multi-scale 3$\times$ training schedule.
This is because the papers~\cite{solo,solov2} of SOLO and SOLOv2 only report the results of multi-scale 6$\times$ schedule,
and the AP$_{s}$, AP$_{m}$, and AP$_{l}$ of CondInst~\cite{tian2020conditional} are calculated based on the areas of bounding boxes rather than instance masks due to implementation bug.
The performance of TensorMask~\cite{chen2019tensormask} is reported from Table 3 of the paper.
The results in Table 2 show that \methodname~obtains better AP$_m$ and AP$_l$ but lower AP$_s$ than Cascade Mask R-CNN.
We hypothesize this is because Cascade Mask R-CNN rescales the regions of small, medium, and large objects to a similar scale of 28 $\times$ 28, and predicts masks on that scale.
On the contrary, \methodname~predicts all the masks on a high-resolution feature map.

\myparagraph{Inference Speed.}
We use frames per second (FPS) to benchmark the inference speed of the models.
Specifically, we benchmark SOLO~\cite{solo}, SOLOv2~\cite{solov2}, CondInst~\cite{tian2020conditional}, Mask R-CNN~\cite{mask_rcnn}, Cascade Mask R-CNN~\cite{cascade_rcnn} and \methodname~with an NVIDIA V100 GPU.
We calculate the pure inference speed of the model without counting in the data loading time, because the latency of data loading depends on the storage system of the testing machine and can vary in different environments.
The reported FPS is an average FPS obtained in three runs, where each run measures the FPS of a model through 400 iterations~\cite{wu2019detectron2, mmdetection}.
Note that the inference speed of these models may be updated due to better implementation and specific optimizations.
So we present them in Table 2 only to verify that \methodname~is fast and effective.

\subsection{Semantic Segmentation}
In Table 3b of the main text, we compare \methodname~on UperNet~\cite{upernet} using Swin Transformer~\cite{SwinTransformer} with the previous state-of-the-art obtained by Swin Transformer~\cite{SwinTransformer}.
We first directly test the model in the last row of Table 3a of the main text (52.0 mIoU) with test-time augmentation and obtain 53.3 mIoU, which is on-par with the current state-of-the-art result (53.5 mIoU).
Then we follow the setting in Swin Transformer~\cite{SwinTransformer} to train the model with larger scale, which resize the long edge and short edge of images to 2048 and 640, respectively, during training and testing.
The model finally obtains 54.3 mIoU on the validation set, which achieves new state-of-the-art performance on ADE20K.

\begin{figure}[t]
    \centering\includegraphics[width=\linewidth]{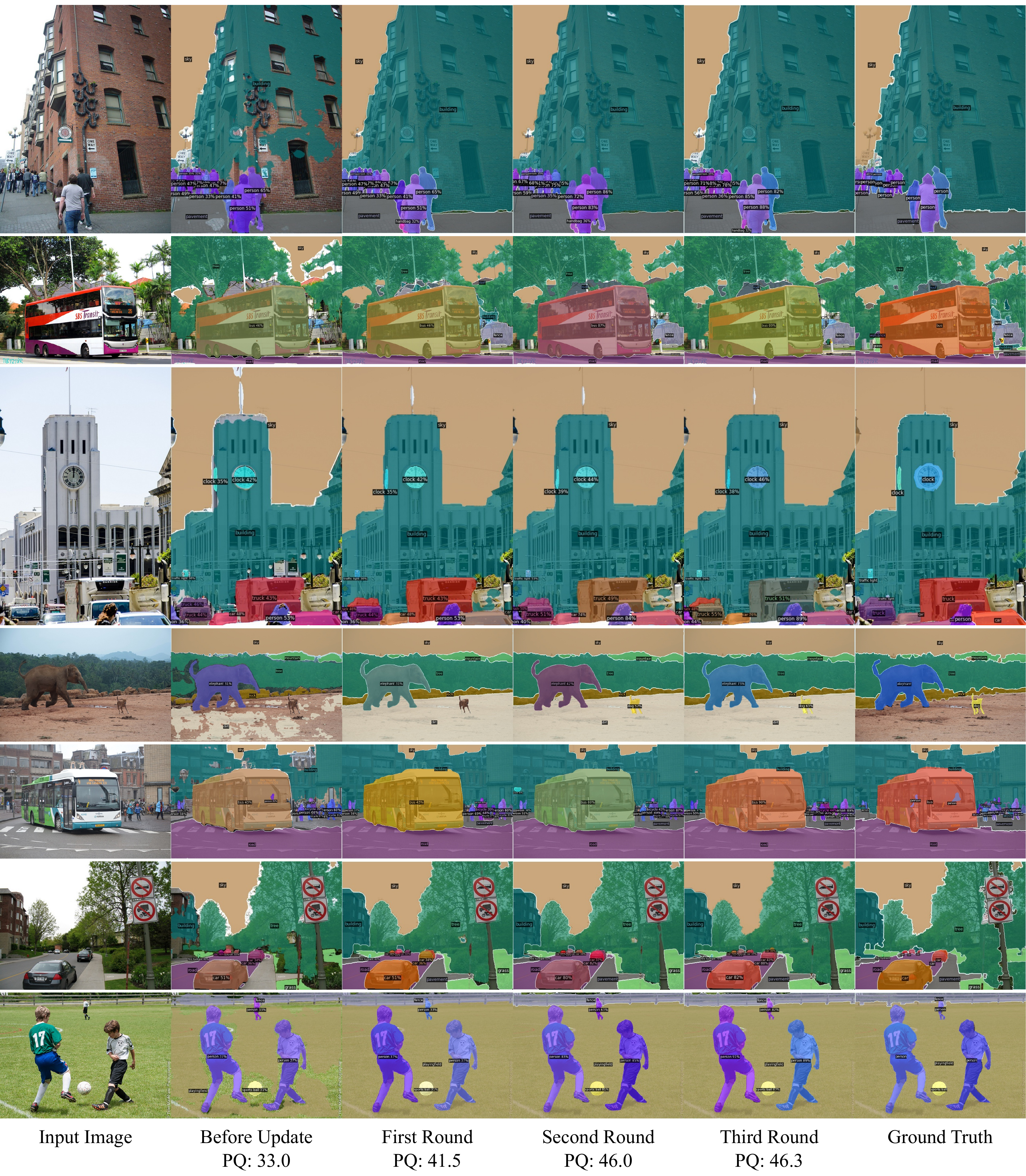}
	\vspace{-12pt}

    \vspace{-6pt}
    \caption{\small{
		Changes of masks before and after kernel updates of a similar model.
        PQ of each stage on the \texttt{val} split is also reported.
        Best viewed in color with zoom in.
	}}\label{fig:mask_preds_stages}
\end{figure}

\begin{figure}[t]
    \centering
    \begin{minipage}[b]{\linewidth}
        \centering\includegraphics[width=\linewidth]{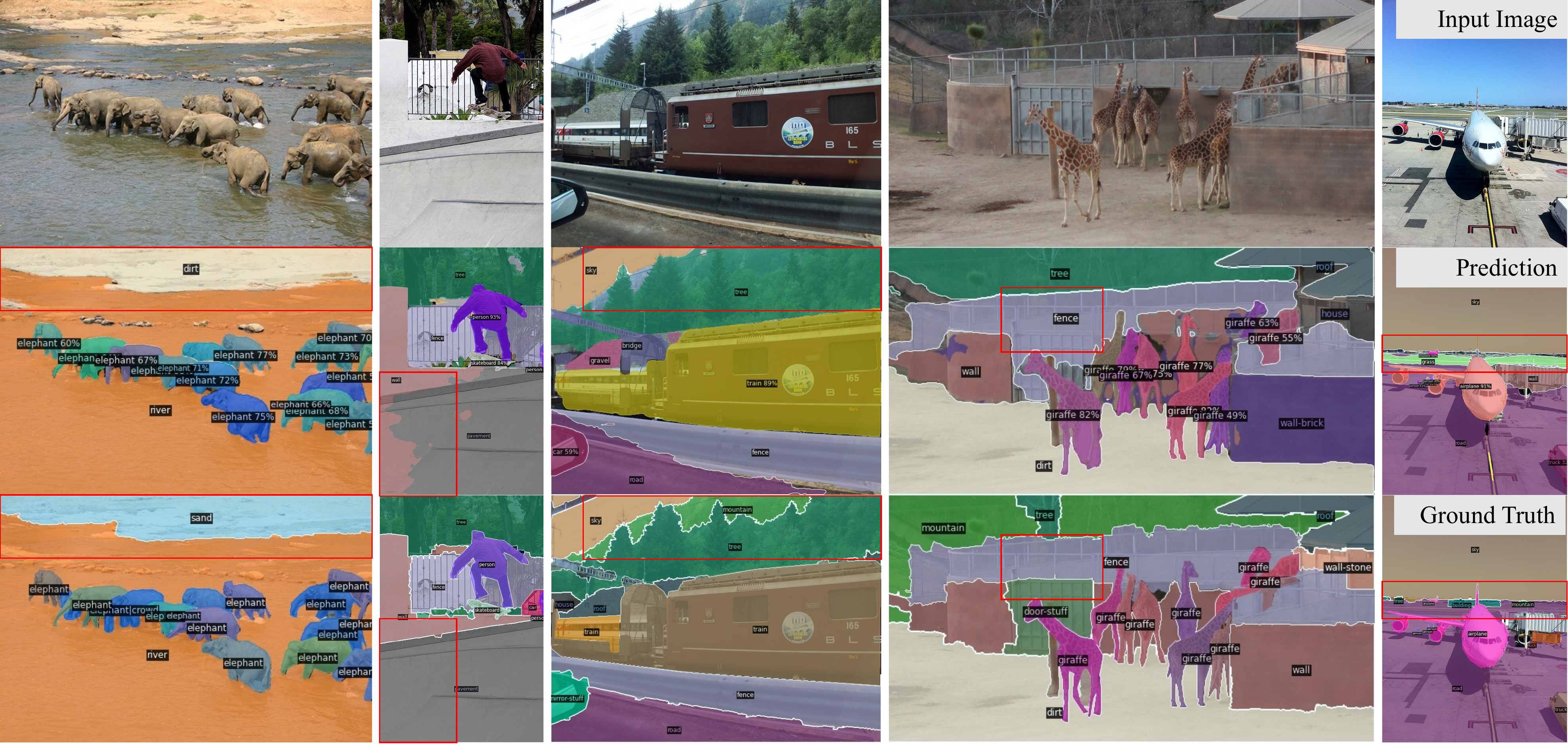}
	    \vspace{-12pt}
	    \subcaption{\footnotesize{
                Misclassification and inaccurate boundaries between contents that share similar texture appearance.
	    	}}\label{fig:mask_preds_texture}
    \end{minipage}
    \vspace{3pt}
    \begin{minipage}[b]{\linewidth}
        \centering\includegraphics[width=\linewidth]{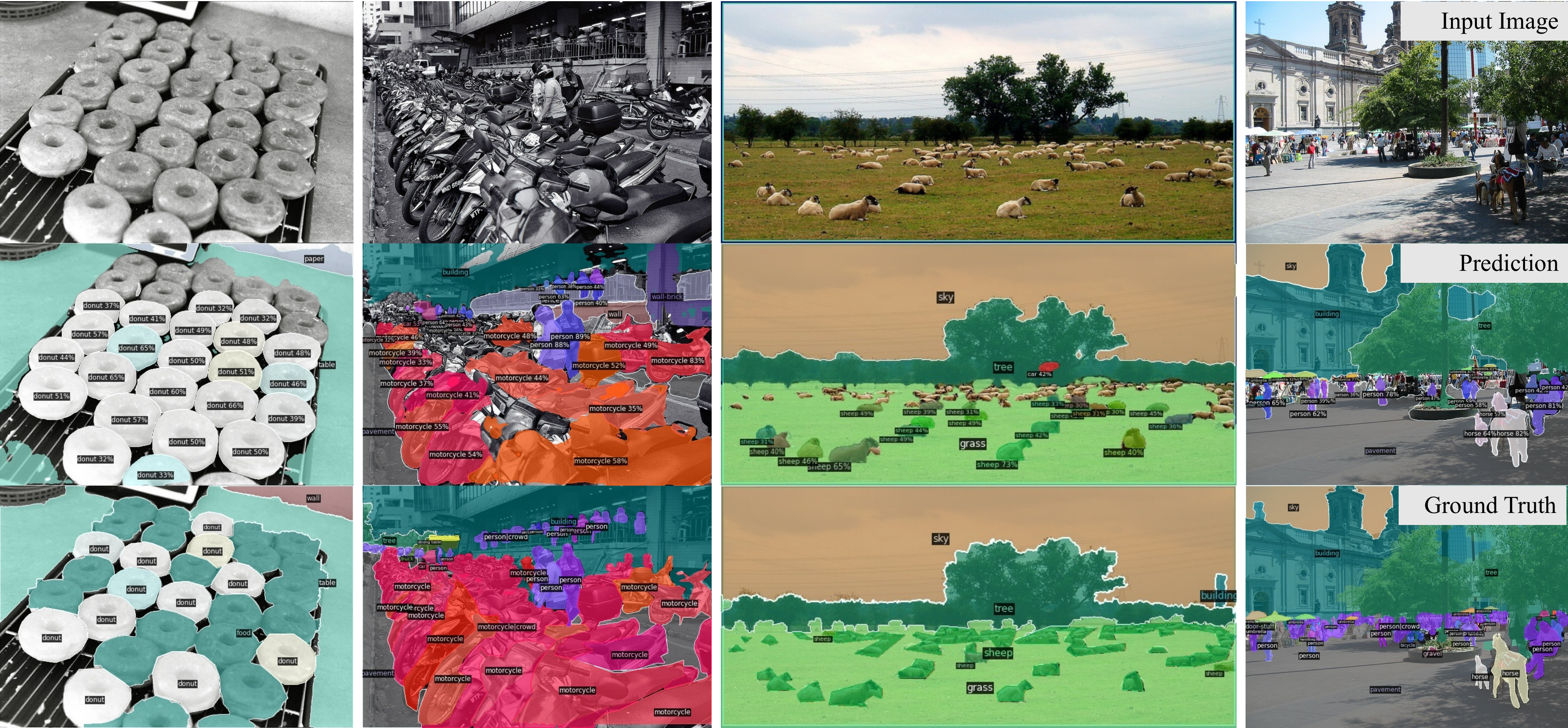}
	    \vspace{-12pt}
	    \subcaption{\footnotesize{
	    		Mask prediction on the images that contain crowded instances.
	    	}}\label{fig:mask_preds_crowd}
    \end{minipage}
    \vspace{-18pt}
    \caption{\small{
		Failure modes of \methodname. Best viewed in color with zoom in.
	}}\label{fig:mask_preds_failure}
\end{figure}

\vspace{-1mm}\section{Visual Analysis}\label{sec:appendix:analysis}\vspace{-1mm}

\myparagraph{Masks Refined through Kernel Update.}
We analyze how the mask predictions change before and after each round of kernel update as shown in Figure~\ref{fig:mask_preds_stages}.
The static kernels have difficulties in handling the boundaries between masks, and the mask prediction cannot cover the whole image.
The mask boundaries are gradually refined and the empty holes in big masks are finally filled through kernel updates.
Notably, the mask predictions after the second and the third rounds look very similar, which means the discriminative capabilities of kernels start to saturate
after the second round kernel update.
The visual analysis is consistent with the evaluation metrics of a similar model on the \texttt{val} split,
where the static kernels before kernel update only achieve 33.0 PQ, and the dynamic kernels after the first update obtain 41.0 PQ.
The dynamic kernels after the second and the third rounds obtain 46.0 PQ and 46.3 PQ, respectively.

\myparagraph{Failure Cases.}
We also analyze the failure cases and find two typical failure modes of \methodname.
First, for the contents that have very similar texture appearance, \methodname~sometimes have difficulties to distinguish them from each other and results in inaccurate mask boundaries and misclassification of contents.
Second, as shown in Figure~\ref{fig:mask_preds_crowd}, in crowded scenarios, it is also challenging for \methodname~to recognize and segment all the instances given limited number of instance kernels.

\vspace{-1mm}\section{Broader Impact}\label{sec:appendix:impact}\vspace{-1mm}
Simplicity and effectiveness are two significant properties pursued by computer vision algorithms.
Our work pushes the boundary of segmentation algorithms through these two aspects by providing a unified perspective that tackles semantic, instance, and panoptic segmentation tasks consistently.
The work could also ease and accelerate the model production and deployment in real-world applications, such as in autonomous driving, robotics, and mobile phones.
The model with higher accuracy proposed in this work could also improve the safety of its related applications.
However, due to limited resources, we do not evaluate the robustness of the proposed method on corrupted images and adversarial attacks.
Therefore, the safety of the applications using this work may not be guaranteed.
To mitigate that, we plan to analyze and improve the robustness of models in the future research.



{\small
  \bibliographystyle{ieee_fullname}
  \bibliography{sections/mainbib}

\begin{thebibliography}{10}\itemsep=-1pt

\bibitem{deep_watershed}
Min Bai and Raquel Urtasun.
\newblock Deep watershed transform for instance segmentation.
\newblock In {\em CVPR}, 2017.

\bibitem{YOLACT}
Daniel Bolya, Chong Zhou, Fanyi Xiao, and Yong~Jae Lee.
\newblock {YOLACT: Real}-time instance segmentation.
\newblock In {\em ICCV}, 2019.

\bibitem{cascade_rcnn}
Zhaowei Cai and Nuno Vasconcelos.
\newblock Cascade {R-CNN:} {Delving} into high quality object detection.
\newblock In {\em CVPR}, 2018.

\bibitem{DETR}
Nicolas Carion, Francisco Massa, Gabriel Synnaeve, Nicolas Usunier, Alexander
  Kirillov, and Sergey Zagoruyko.
\newblock End-to-end object detection with transformers.
\newblock In {\em ECCV}, 2020.

\bibitem{htc}
Kai Chen, Jiangmiao Pang, Jiaqi Wang, Yu Xiong, Xiaoxiao Li, Shuyang Sun,
  Wansen Feng, Ziwei Liu, Jianping Shi, Wanli Ouyang, Chen~Change Loy, and
  Dahua Lin.
\newblock Hybrid task cascade for instance segmentation.
\newblock In {\em CVPR}, 2019.

\bibitem{mmdetection}
Kai Chen, Jiaqi Wang, Jiangmiao Pang, Yuhang Cao, Yu Xiong, Xiaoxiao Li,
  Shuyang Sun, Wansen Feng, Ziwei Liu, Jiarui Xu, Zheng Zhang, Dazhi Cheng,
  Chenchen Zhu, Tianheng Cheng, Qijie Zhao, Buyu Li, Xin Lu, Rui Zhu, Yue Wu,
  Jifeng Dai, Jingdong Wang, Jianping Shi, Wanli Ouyang, Chen~Change Loy, and
  Dahua Lin.
\newblock {MMDetection}: Open mmlab detection toolbox and benchmark.
\newblock {\em arXiv preprint arXiv:1906.07155}, 2019.

\bibitem{deeplab}
Liang{-}Chieh Chen, George Papandreou, Iasonas Kokkinos, Kevin Murphy, and
  Alan~L. Yuille.
\newblock {DeepLab: Semantic} image segmentation with deep convolutional nets,
  atrous convolution, and fully connected crfs.
\newblock {\em TPAMI}, 2018.

\bibitem{deeplabv3}
Liang-Chieh Chen, George Papandreou, Florian Schroff, and Hartwig Adam.
\newblock Rethinking atrous convolution for semantic image segmentation.
\newblock {\em arXiv preprint arXiv:1706.05587}, 2017.

\bibitem{deeplabv3plus}
Liang-Chieh Chen, Yukun Zhu, George Papandreou, Florian Schroff, and Hartwig
  Adam.
\newblock Encoder-decoder with atrous separable convolution for semantic image
  segmentation.
\newblock In {\em ECCV}, 2018.

\bibitem{chen2019tensormask}
Xinlei Chen, Ross Girshick, Kaiming He, and Piotr Dollár.
\newblock Tensormask: A foundation for dense object segmentation.
\newblock 2019.

\bibitem{panoptic_deeplab}
Bowen Cheng, Maxwell~D. Collins, Yukun Zhu, Ting Liu, Thomas~S. Huang, Hartwig
  Adam, and Liang{-}Chieh Chen.
\newblock {Panoptic-DeepLab: A} simple, strong, and fast baseline for bottom-up
  panoptic segmentation.
\newblock In {\em CVPR}, 2020.

\bibitem{cheng2021maskformer}
Bowen Cheng, Alexander~G. Schwing, and Alexander Kirillov.
\newblock Per-pixel classification is not all you need for semantic
  segmentation.
\newblock {\em CoRR}, abs/2107.06278, 2021.

\bibitem{mmseg2020}
MMSegmentation Contributors.
\newblock {MMSegmentation}: Openmmlab semantic segmentation toolbox and
  benchmark.
\newblock \url{https://github.com/open-mmlab/mmsegmentation}, 2020.

\bibitem{Cityscapes}
Marius Cordts, Mohamed Omran, Sebastian Ramos, Timo Rehfeld, Markus Enzweiler,
  Rodrigo Benenson, Uwe Franke, Stefan Roth, and Bernt Schiele.
\newblock The cityscapes dataset for semantic urban scene understanding.
\newblock In {\em CVPR}, 2016.

\bibitem{Dai_2016}
Jifeng Dai, Kaiming He, and Jian Sun.
\newblock Instance-aware semantic segmentation via multi-task network cascades.
\newblock {\em CVPR}, 2016.

\bibitem{dcn}
Jifeng Dai, Haozhi Qi, Yuwen Xiong, Yi Li, Guodong Zhang, Han Hu, and Yichen
  Wei.
\newblock Deformable convolutional networks.
\newblock In {\em ICCV}, 2017.

\bibitem{VIT}
Alexey Dosovitskiy, Lucas Beyer, Alexander Kolesnikov, Dirk Weissenborn,
  Xiaohua Zhai, Thomas Unterthiner, Mostafa Dehghani, Matthias Minderer, Georg
  Heigold, Sylvain Gelly, Jakob Uszkoreit, and Neil Houlsby.
\newblock An image is worth 16x16 words: Transformers for image recognition at
  scale.
\newblock In {\em ICLR}, 2021.

\bibitem{deformable_kernel}
Hang Gao, Xizhou Zhu, Stephen Lin, and Jifeng Dai.
\newblock {Deformable Kernels: Adapting} effective receptive fields for object
  deformation.
\newblock In {\em ICLR}, 2020.

\bibitem{Gould_2009}
Stephen Gould, Tianshi Gao, and Daphne Koller.
\newblock Region-based segmentation and object detection.
\newblock In Yoshua Bengio, Dale Schuurmans, John~D. Lafferty, Christopher
  K.~I. Williams, and Aron Culotta, editors, {\em NeurIPS}, 2009.

\bibitem{DMNet}
Junjun He, Zhongying Deng, and Yu Qiao.
\newblock Dynamic multi-scale filters for semantic segmentation.
\newblock In {\em ICCV}, 2019.

\bibitem{He_2019_ICCV}
Kaiming He, Ross Girshick, and Piotr Dollar.
\newblock Rethinking {ImageNet} pre-training.
\newblock In {\em ICCV}, 2019.

\bibitem{mask_rcnn}
Kaiming He, Georgia Gkioxari, Piotr Doll{\'{a}}r, and Ross~B. Girshick.
\newblock Mask {R-CNN}.
\newblock In {\em ICCV}, 2017.

\bibitem{He_2016}
Kaiming He, Xiangyu Zhang, Shaoqing Ren, and Jian Sun.
\newblock Deep residual learning for image recognition.
\newblock In {\em CVPR}, 2016.

\bibitem{ms_rcnn}
Zhaojin Huang, Lichao Huang, Yongchao Gong, Chang Huang, and Xinggang Wang.
\newblock {Mask Scoring R-CNN}.
\newblock In {\em CVPR}, 2019.

\bibitem{STN}
Max Jaderberg, Karen Simonyan, Andrew Zisserman, and Koray Kavukcuoglu.
\newblock Spatial transformer networks.
\newblock In {\em NeurIPS}, 2015.

\bibitem{DFN}
Xu Jia, Bert~De Brabandere, Tinne Tuytelaars, and Luc~Van Gool.
\newblock Dynamic filter networks.
\newblock In {\em NeurIPS}, 2016.

\bibitem{BCNet}
Lei Ke, Yu{-}Wing Tai, and Chi{-}Keung Tang.
\newblock Deep occlusion-aware instance segmentation with overlapping bilayers.
\newblock In {\em CVPR}, 2021.

\bibitem{panoptic_fpn}
Alexander Kirillov, Ross~B. Girshick, Kaiming He, and Piotr Doll{\'{a}}r.
\newblock Panoptic feature pyramid networks.
\newblock In {\em CVPR}, 2019.

\bibitem{panoptic}
Alexander Kirillov, Kaiming He, Ross Girshick, Carsten Rother, and Piotr
  Doll{\'a}r.
\newblock Panoptic segmentation.
\newblock In {\em CVPR}, 2019.

\bibitem{instance_cut}
Alexander Kirillov, Evgeny Levinkov, Bjoern Andres, Bogdan Savchynskyy, and
  Carsten Rother.
\newblock Instancecut: From edges to instances with multicut.
\newblock In {\em CVPR}, 2017.

\bibitem{unifying_CVPR2020}
Qizhu Li, Xiaojuan Qi, and Philip H.~S. Torr.
\newblock Unifying training and inference for panoptic segmentation.
\newblock In {\em CVPR}, 2020.

\bibitem{emanet}
Xia Li, Zhisheng Zhong, Jianlong Wu, Yibo Yang, Zhouchen Lin, and Hong Liu.
\newblock Expectation-maximization attention networks for semantic
  segmentation.
\newblock In {\em ICCV}, 2019.

\bibitem{FCIS}
Yi Li, Haozhi Qi, Jifeng Dai, Xiangyang Ji, and Yichen Wei.
\newblock Fully convolutional instance-aware semantic segmentation.
\newblock {\em CVPR}, 2017.

\bibitem{panoptic_fcn}
Yanwei Li, Hengshuang Zhao, Xiaojuan Qi, Liwei Wang, Zeming Li, Jian Sun, and
  Jiaya Jia.
\newblock Fully convolutional networks for panoptic segmentation.
\newblock In {\em CVPR}, 2021.

\bibitem{panoptic_segformer}
Zhiqi Li, Wenhai Wang, Enze Xie, Zhiding Yu, Anima Anandkumar, Jose~M. Alvarez,
  Tong Lu, and Ping Luo.
\newblock Panoptic segformer.
\newblock {\em CoRR}, abs/2109.03814, 2021.

\bibitem{lin2017_fpn}
Tsung{-}Yi Lin, Piotr Doll{\'{a}}r, Ross~B. Girshick, Kaiming He, Bharath
  Hariharan, and Serge~J. Belongie.
\newblock Feature pyramid networks for object detection.
\newblock In {\em CVPR}, 2017.

\bibitem{lin2017_focal}
Tsung{-}Yi Lin, Priya Goyal, Ross~B. Girshick, Kaiming He, and Piotr
  Doll{\'{a}}r.
\newblock Focal loss for dense object detection.
\newblock In {\em ICCV}, 2017.

\bibitem{lin2014coco}
Tsung-Yi Lin, Michael Maire, Serge Belongie, James Hays, Pietro Perona, Deva
  Ramanan, Piotr Doll{\'a}r, and C~Lawrence Zitnick.
\newblock Microsoft {COCO}: Common objects in context.
\newblock In {\em ECCV}, 2014.

\bibitem{SwinTransformer}
Ze Liu, Yutong Lin, Yue Cao, Han Hu, Yixuan Wei, Zheng Zhang, Stephen Lin, and
  Baining Guo.
\newblock {Swin Transformer: Hierarchical} vision transformer using shifted
  windows.
\newblock {\em arXiv preprint arXiv:2103.14030}, 2021.

\bibitem{FCN}
Jonathan Long, Evan Shelhamer, and Trevor Darrell.
\newblock Fully convolutional networks for semantic segmentation.
\newblock In {\em CVPR}, 2015.

\bibitem{ADAMW}
Ilya Loshchilov and Frank Hutter.
\newblock Decoupled weight decay regularization.
\newblock In {\em ICLR}, 2019.

\bibitem{dice_loss}
Fausto Milletari, Nassir Navab, and Seyed{-}Ahmad Ahmadi.
\newblock {V-Net: Fully} convolutional neural networks for volumetric medical
  image segmentation.
\newblock In {\em 3DV}, 2016.

\bibitem{NevenBPG19}
Davy Neven, Bert~De Brabandere, Marc Proesmans, and Luc~Van Gool.
\newblock Instance segmentation by jointly optimizing spatial embeddings and
  clustering bandwidth.
\newblock In {\em CVPR}, 2019.

\bibitem{associative_embedding}
Alejandro Newell, Zhiao Huang, and Jia Deng.
\newblock {Associative Embedding: End}-to-end learning for joint detection and
  grouping.
\newblock In {\em NeurIPS}, 2017.

\bibitem{deep_snake}
Sida Peng, Wen Jiang, Huaijin Pi, Xiuli Li, Hujun Bao, and Xiaowei Zhou.
\newblock Deep snake for real-time instance segmentation.
\newblock In {\em CVPR}, 2020.

\bibitem{ren2015faster}
Shaoqing Ren, Kaiming He, Ross Girshick, and Jian Sun.
\newblock Faster {R-CNN}: Towards real-time object detection with region
  proposal networks.
\newblock In {\em NeurIPS}, 2015.

\bibitem{set_prediction_loss}
Russell Stewart, Mykhaylo Andriluka, and Andrew~Y. Ng.
\newblock End-to-end people detection in crowded scenes.
\newblock In {\em CVPR}, 2016.

\bibitem{szeliski2010computer}
Richard Szeliski.
\newblock {\em Computer vision: algorithms and applications}.
\newblock Springer Science \& Business Media, 2010.

\bibitem{tian2019adelaidet}
Zhi Tian, Hao Chen, Xinlong Wang, Yuliang Liu, and Chunhua Shen.
\newblock {AdelaiDet}: A toolbox for instance-level recognition tasks.
\newblock \url{https://git.io/adelaidet}, 2019.

\bibitem{tian2020conditional}
Zhi Tian, Chunhua Shen, and Hao Chen.
\newblock Conditional convolutions for instance segmentation.
\newblock In {\em ECCV}, 2020.

\bibitem{tu_2005}
Zhuowen Tu, Xiangrong Chen, Alan~L. Yuille, and Song~Chun Zhu.
\newblock Image parsing: Unifying segmentation, detection, and recognition.
\newblock {\em IJCV}, 2005.

\bibitem{transformer}
Ashish Vaswani, Noam Shazeer, Niki Parmar, Jakob Uszkoreit, Llion Jones,
  Aidan~N. Gomez, Lukasz Kaiser, and Illia Polosukhin.
\newblock Attention is all you need.
\newblock In {\em NeurIPS}, 2017.

\bibitem{max_deeplab}
Huiyu Wang, Yukun Zhu, Hartwig Adam, Alan~L. Yuille, and Liang{-}Chieh Chen.
\newblock Max-deeplab: End-to-end panoptic segmentation with mask transformers.
\newblock {\em CoRR}, abs/2012.00759, 2020.

\bibitem{PVTv2}
Wenhai Wang, Enze Xie, Xiang Li, Deng{-}Ping Fan, Kaitao Song, Ding Liang, Tong
  Lu, Ping Luo, and Ling Shao.
\newblock {PVTv2: Improved} baselines with pyramid vision transformer.
\newblock {\em CoRR}, abs/2106.13797, 2021.

\bibitem{solo}
Xinlong Wang, Tao Kong, Chunhua Shen, Yuning Jiang, and Lei Li.
\newblock {SOLO: Segmenting} objects by locations.
\newblock In {\em ECCV}, 2020.

\bibitem{solov2}
Xinlong Wang, Rufeng Zhang, Tao Kong, Lei Li, and Chunhua Shen.
\newblock {SOLOv2: Dynamic} and fast instance segmentation.
\newblock {\em NeurIPS}, 2020.

\bibitem{dynamic_filtering_eccv2018}
Jialin Wu, Dai Li, Yu Yang, Chandrajit Bajaj, and Xiangyang Ji.
\newblock Dynamic filtering with large sampling field for convnets.
\newblock In {\em ECCV}, 2018.

\bibitem{wu2019detectron2}
Yuxin Wu, Alexander Kirillov, Francisco Massa, Wan-Yen Lo, and Ross Girshick.
\newblock Detectron2.
\newblock \url{https://github.com/facebookresearch/detectron2}, 2019.

\bibitem{upernet}
Tete Xiao, Yingcheng Liu, Bolei Zhou, Yuning Jiang, and Jian Sun.
\newblock Unified perceptual parsing for scene understanding.
\newblock In {\em ECCV}, 2018.

\bibitem{polarmask}
Enze Xie, Peize Sun, Xiaoge Song, Wenhai Wang, Xuebo Liu, Ding Liang, Chunhua
  Shen, and Ping Luo.
\newblock {PolarMask: Single} shot instance segmentation with polar
  representation.
\newblock In {\em CVPR}, 2020.

\bibitem{UPSNet}
Yuwen Xiong, Renjie Liao, Hengshuang Zhao, Rui Hu, Min Bai, Ersin Yumer, and
  Raquel Urtasun.
\newblock {UPSNet: A} unified panoptic segmentation network.
\newblock In {\em CVPR}, 2019.

\bibitem{ese_seg}
Wenqiang Xu, Haiyang Wang, Fubo Qi, and Cewu Lu.
\newblock Explicit shape encoding for real-time instance segmentation.
\newblock In {\em ICCV}, 2019.

\bibitem{deeperlab}
Tien{-}Ju Yang, Maxwell~D. Collins, Yukun Zhu, Jyh{-}Jing Hwang, Ting Liu, Xiao
  Zhang, Vivienne Sze, George Papandreou, and Liang{-}Chieh Chen.
\newblock {DeeperLab: Single}-shot image parser.
\newblock {\em CoRR}, abs/1902.05093, 2019.

\bibitem{holistic_scene_understanding}
Jian Yao, Sanja Fidler, and Raquel Urtasun.
\newblock Describing the scene as a whole: Joint object detection, scene
  classification and semantic segmentation.
\newblock In {\em CVPR}, 2012.

\bibitem{DNL}
Minghao Yin, Zhuliang Yao, Yue Cao, Xiu Li, Zheng Zhang, Stephen Lin, and Han
  Hu.
\newblock Disentangled non-local neural networks.
\newblock In {\em ECCV}, 2020.

\bibitem{OCRNet}
Yuhui Yuan, Xilin Chen, and Jingdong Wang.
\newblock Object-contextual representations for semantic segmentation.
\newblock In {\em ECCV}, 2020.

\bibitem{encnet}
Hang Zhang, Kristin Dana, Jianping Shi, Zhongyue Zhang, Xiaogang Wang, Ambrish
  Tyagi, and Amit Agrawal.
\newblock Context encoding for semantic segmentation.
\newblock In {\em CVPR}, June 2018.

\bibitem{resnest}
Hang Zhang, Chongruo Wu, Zhongyue Zhang, Yi Zhu, Zhi Zhang, Haibin Lin, Yue
  Sun, Tong He, Jonas Muller, R. Manmatha, Mu Li, and Alexander Smola.
\newblock Resnest: Split-attention networks.
\newblock {\em arXiv preprint arXiv:2004.08955}, 2020.

\bibitem{PSPNet}
Hengshuang Zhao, Jianping Shi, Xiaojuan Qi, Xiaogang Wang, and Jiaya Jia.
\newblock Pyramid scene parsing network.
\newblock In {\em CVPR}, 2017.

\bibitem{psanet}
Hengshuang Zhao, Yi Zhang, Shu Liu, Jianping Shi, Chen Change~Loy, Dahua Lin,
  and Jiaya Jia.
\newblock {PSANet: Point-wise} spatial attention network for scene parsing.
\newblock In {\em ECCV}, 2018.

\bibitem{SETR}
Sixiao Zheng, Jiachen Lu, Hengshuang Zhao, Xiatian Zhu, Zekun Luo, Yabiao Wang,
  Yanwei Fu, Jianfeng Feng, Tao Xiang, Philip~H.S. Torr, and Li Zhang.
\newblock Rethinking semantic segmentation from a sequence-to-sequence
  perspective with transformers.
\newblock In {\em CVPR}, 2021.

\bibitem{ADE20K}
Bolei Zhou, Hang Zhao, Xavier Puig, Tete Xiao, Sanja Fidler, Adela Barriuso,
  and Antonio Torralba.
\newblock Semantic understanding of scenes through the {ADE20K} dataset.
\newblock {\em IJCV}, 2019.

\bibitem{dcnv2}
Xizhou Zhu, Han Hu, Stephen Lin, and Jifeng Dai.
\newblock {Deformable ConvNets V2: More} deformable, better results.
\newblock In {\em CVPR}, 2019.

\bibitem{DeformableDETR}
Xizhou Zhu, Weijie Su, Lewei Lu, Bin Li, Xiaogang Wang, and Jifeng Dai.
\newblock Deformable {DETR:} deformable transformers for end-to-end object
  detection.
\newblock {\em CoRR}, abs/2010.04159, 2020.

\end{thebibliography}
}


\end{document}